\documentclass[journal]{IEEEtran}
\ifCLASSINFOpdf
\else
\fi
\usepackage{color}
\usepackage{amsmath,amssymb} 
\usepackage{amsthm}
\usepackage{amsmath}
\usepackage[english]{babel}
\usepackage[utf8]{inputenc}

\usepackage{epsfig}
\usepackage{graphicx}
\usepackage{caption}
\usepackage{subcaption}

\usepackage{cite}
\usepackage{siunitx}
\usepackage{enumerate}
\usepackage{siunitx}
\usepackage{enumitem}

\usepackage{ragged2e}
\usepackage{xcolor}

\usepackage[font={small}]{caption}
\usepackage{multirow}
\usepackage{algpseudocode}
\usepackage{algorithm}

\usepackage{flushend,cuted}
\usepackage{verbatim}

\usepackage{footmisc}

\newcolumntype{L}[1]{>{\raggedright\let\newline\\\arraybackslash\hspace{0pt}}m{#1}}
\newcolumntype{C}[1]{>{\centering\let\newline\\\arraybackslash\hspace{0pt}}m{#1}}
\newcolumntype{R}[1]{>{\raggedleft\let\newline\\\arraybackslash\hspace{0pt}}m{#1}}

\begin{document}
\title{Ground Plane based Absolute Scale Estimation for Monocular Visual Odometry}
%
%
%

\author{Dingfu~Zhou,
Yuchao~Dai,~\IEEEmembership{Member,~IEEE,}
and~Hongdong~Li,~\IEEEmembership{Member,~IEEE}
\thanks{

Dingfu Zhou is with Baidu Inc., Beijing, China and National Engineering Laboratory of Deep Learning Technology and Application, China. 

Yuchao Dai is with School of Electronics and Information, Northwestern Polytechnical University, Xi’an, China. Yuchao Dai is the corresponding author (daiyuchao@gmail.com).

Hongdong Li is with Research School of Engineering, the Australian National University, Australia and Australia Centre for Robotic Vision.}
}

\markboth{IEEE Transactions on xxxx xxxx }
{Zhou \MakeLowercase{\textit{et al.}}: Ground Plane based Absolute Scale Estimation for Monocular Visual Odometry}

\maketitle

	\maketitle
	
\begin{abstract}
Recovering absolute metric scale from a monocular camera is a challenging but highly desirable problem for monocular camera-based systems. By using different kinds of cues, various approaches have been proposed for scale estimation, such as camera height, object size etc. In this paper, firstly, we summarize different kinds of scale estimation approaches. Then, we propose a robust divide and conquer absolute scale estimation method based on the ground plane and camera height by analyzing the advantages and disadvantages of different approaches. By using the estimated scale, an effective scale correction strategy has been proposed to reduce the scale drift during the Monocular Visual Odometry (VO) estimation process. Finally, the effectiveness and robustness of the proposed method have been verified on both public and self-collected image sequences. 
\end{abstract}
	
	
	%

	
	\section{Introduction\label{sec:introduction}}
	%
	%
	%
	%

\IEEEPARstart{V}{ision} based Structure-from-Motion (SfM), Visual Odometry (VO) and Simultaneous Localization And Mapping (SLAM) play important roles in advanced driver assistance and autonomous driving systems. Compared with other active sensors, such as Light Detection And Ranging (Lidar), vision-based systems have several advantages: first, the camera sensor is very cheap; second, cameras can provide color, semantic and geometric information, which are important for scene understanding; finally, cameras which are passive sensors need less power consumption than the active sensors. Different from stereo or multi-cameras vision systems, monocular camera system \cite{mirabdollah2015fast}, \cite{forster2014svo}, \cite{mur2015orb} is a very attractive option for real-world applications due to its own merits: a single camera is easy to be mounted on the vehicle and it is also free from the burden of multi-camera calibration. Furthermore, a single fish-eye camera can also provide a relatively large field of view as stereo rigs.

However, all monocular camera-based systems suffer from one drawback, which is called as similarity ambiguity \cite{hartley2003multiple}. In other words, from only monocular camera images (a pair of view or number of views), we cannot recover the 3D structure and camera motion with absolute metric information. Without prior knowledge, we don't know whether the reconstructed scene is a real or just an artificial model. To recover the absolute scale, at least one piece of metric information is required. This cue may come from prior scene knowledge (e.g., camera height, object size, vehicle speed, baseline of stereo camera etc.) or from other sensors, such as IMU (Inertial Measurement Unit) or GPS (Global Positioning System) etc. If two cameras are provided, i.e. forming a binocular system, the absolute scale can be recovered based on the baseline of two cameras. Alternatively, the absolute scale can also be recovered by using IMU, GPS \cite{hu2004real,shepard2014high} or wheel odometry \cite{zhang2014robust}, etc. Furthermore, prior geometric knowledge, such as the camera height, object size, has also been employed for absolute scale recovery. 
	
Scale-drift (accumulative error) is another big problem in VO and SLAM systems. This error usually comes from the procedure of features extraction, feature tracking, 3D reconstruction and relative pose estimation. In addition, this kind of error is unavoidable in the existing SLAM systems because 3D camera trajectory and environment map are reconstructed incrementally frame by frame. To overcome this kind of error, many approaches have been proposed. To reduce the uncertainty generated during the initialization and tracking process, a concept of inverse depth parametrization \cite{civera2008inverse} has been proposed for monocular SLAM system. To reduce the feature matching and tracking error, a feature descriptor called ``synthetic basis descriptor'' \cite{desai2016visual} was developped to match features across different views. In addition, a sliding window strategy is also applied for extending feature transformation into subsequent frames to overcome the limitation of the short baseline nature of VO.
	
Tracking features in multi-frames have been proved to be an effective way for reducing the scale drift in VO \cite{badino2013visual}, e.g., local Bundle Adjustment (BA) technique. Based on BA, the rotation and translation error can be minimized across multi-views. Although BA technique is effective in small-scale environment, the drift is also serious after a long distance driving in real-world traffic scenarios.
	
Besides BA, loop closure is another effective strategy for scale drift reduction which relies on place recognition technique. If a loop is detected, the current camera will be forced to relocate to a previous prior location. Usually, this procedure should be executed carefully because an incorrect loop detection may result in a crash of the whole system. In addition, in the real autonomous driving scenarios, the loop does not always exist. Especially in the high-way scenarios, the drift becomes more serious because all the features can only survive in a few frames due to the high speed of the vehicle.

This article is organized as follows: first, Section \ref{sec:related_works} discusses related works for scale estimation and correction. Next, camera height based scale estimation approaches are introduced in Section \ref{sec:Camera_height_based_scale_estimation}. Then the proposed robust divide and conquer scale estimation and correction method is presented in Section \ref{sec:Proposed_method} and \ref{sec:Scale_correction} respectively. We evaluated the proposed method on both synthetic and real KITTI image sequences in Section \ref{sec:Experimental_results}. Finally, our paper ends with a conclusion and discussion of potential future works.

	\section{Related Works}\label{sec:related_works}
	VO, SfM and SLAM have been widely researched for more than 30 years. A comprehensive review of these works is beyond the scope of this paper. Some detailed summaries of them can be found in \cite{scaramuzza2011visual,fraundorfer2012visual}, \cite{ozyecsil2017survey} and \cite{fuentes2015visual}. In this section, we only discuss latest works related to our absolute scale estimation problem.
	
	\subsection{Multi-sensor based methods} 
	Recently, Lidar has been widely employed \cite{cole2006using} or fused with cameras \cite{zhang2015visual} for VO and SLAM. Based on the depth sensor, the absolute scale can be recovered directly. However, motion estimation based Lidar point cloud suffers from the so-called motion distortion effect because the range measurements are received at different times during continuous Lidar motion. More seriously, scan matching also fails in some degenerate cases when the scene is dominated by planar areas. A combination of camera and Lidar for motion estimation is also commonly employed to enhance the advantage and avoid the disadvantage of each other.
	
	Inertial measurement unit (IMU) sensors have also been used for VO and SLAM by fusing with cameras \cite{nutzi2011fusion}, \cite{martinelli2012vision}, \cite{lupton2008removing}, \cite{grabe2013comparison}, \cite{mustaniemi2017inertial}, \cite{xiong2017scale}. IMU sensor can give a 3D acceleration of rotation and translation of a moving platform. The platform's position can be obtained by a second order integral. Unfortunately, direct integration of acceleration measurements drifts quadratically in time due to the measurement noise. Conversely, the drift of camera-based VO is relatively small. So intelligent combination of IMU and camera can provide not only scale but also a stable estimation. In general, IMU aided approach can be categorized either as tightly or loosely coupled \cite{tardif2010new} depends on the way of using the IMU in the system. Stereo camera \cite{mur2015orb},\cite{cvivsic2015stereo,buczko2016distinguish,persson2015robust,geiger2011stereoscan} is also an important sensor for VO and SLAM. Based on the baseline, the absolute scale can be recovered easily. By using this information, both the camera motion and the scene structure are reconstructed in metric. A general review of stereo vision based VO can be found in \cite{scaramuzza2011visual,fraundorfer2012visual}.
	
\subsection{Monocular-based methods}
Compared to methods with additional sensors, monocular camera-based approaches are more attractive. In order to estimate the absolute scale, the scene knowledge should be well unitized.
	
\subsubsection{Camera height based methods} Camera height is a commonly used information for absolute scale estimation \cite{mirabdollah2015fast}, \cite{choi2015new}, \cite{scaramuzza2009absolute}, \cite{scaramuzza2008appearance}, \cite{kitt2011monocular}, \cite{lovegrove2011accurate}, \cite{grater2015robust}, \cite{gutierrez2012full},\cite{pereira2017monocular} based on a flat ground plane assumption. In \cite{pereira2017monocular} a multi-attribute cost function has been designed for selecting ground features to estimate the absolute scale. In this paper, a good ground point should be in the center bottom area of the image, present a good image gradient along the epipolar line to enhance matching precision and is expected to be close to the estimated ground.
A multi-modal mechanism of prediction, classification, and correction have been proposed in \cite{fanani2017multimodal}. For robust estimation, the scale correction scheme combines cues from both dense and sparse ground plane estimation. Furthermore, a classification strategy is employed to detect scale outliers based on various features (e.g. moments on residuals). More camera height based method will be detailedly introduced in Section \ref{sec:Camera_height_based_scale_estimation}. 

\subsubsection{Other scene knowledge based methods}
The ground is not always detected in the real scenario. In that case, other scene knowledge has been considered for scale estimation. Botterill et al. proposed to use objects' size to reduce scale drift\cite{botterill2011bag}, \cite{botterill2013correcting}. In these works, an algorithm called SCORE2 (scale correction by object recognition) has been proposed: the objects in surrounding environment are detected and recognized first; then the distribution of objects' size for each class is estimated. Finally, the objects' sizes are used for scale correction when they appear next time.

The detected object's bounding box is also used for scale estimation in \cite{song2015high}. The scale is estimated by minimizing the difference between current detected object's size and prior knowledge. A monocular SLAM system has been developed for indoor robot navigation in \cite{hilsenbeck2012scale}. For scale initialization, a fast and robust method was proposed by using building's geometric properties, e.g., room or corridor's size. A person's height is also used for scale recovering in \cite{lim2015monocular}. First, the true camera's height $h_0$ is estimated by using the 2D pedestrian detection in the image by assuming that the person is in an upright position. Then, the camera height $h_1$ can also be recovered by reconstructing the 3D points on the ground plane via the estimated camera motion (up to a scale). Finally, the absolute scale is the ratio between $h_0$ and $h_1$. In \cite{sucar2017probabilistic}, the global scale of the 3D reconstruction is recovered by a set of pre-defined classes of objects. Other scene knowledge is also used to recover the absolute scale, such as the average pedestrian's height \cite{botterill2013correcting}, vanishing lines \cite{grater2015robust} etc.

\subsubsection{Deep learning based methods} All methods mentioned above explicitly utilize the geometry information for scale estimation. Recently, with the development of deep learning, many approaches have been proposed to learn the camera pose and scale  from image sequences implicitly. In \cite{mohanty2016deepvo} and \cite{konda2015learning}, an end-to-end Convolutional Neural Networks (CNN) based framework has been designed for VO estimation. For VO, brightness constancy between consecutive frames is a common used assumption, however, this doesn't hold in High Dynamic Range (HDR) environment. In \cite{gomez2017learning}, a deep learning based approach has been proposed to obtain enhanced representations of the sequences; then an insertion of Long Short-Term Memory (LSTM) based strategy is applied to obtain temporally consistent sequences. In \cite{wang2017end}, an end-to-end sequence-to-sequence Probabilistic Visual Odometry (ESP-VO) approach has been proposed for monocular VO based on deep recurrent neural networks (RNN). The proposed approach can not only automatically learn effective feature representation, but also implicitly model sequential dynamics and relation for VO with the help of deep RNN. In \cite{costante2018ls} and \cite{zhao2018learning}, two approaches have been proposed to robust estimate the VO by considering the optical flow caused by the camera motion. In \cite{zhao2018learning}, the camera motion has been estimated by using the constraints with depth and optical flow. In \cite{costante2018ls}, a novel network architecture for estimating monocular camera motion which is composed of two branches that jointly learn a latent space representation of the input optical flow field and the camera motion estimate. In \cite{zhan2018unsupervised}, an unsupervised approach has been proposed to recover both depth and camera motion together. During the training process, stereo sequences are used which can provide both spatial (between left-right pairs) and temporal (forward-backward) photometric warp error, and this error constrains the scene depth and camera motion to be estimated in a real-world scale. At test time the framework is able to estimate single view depth and two-view odometry from a monocular sequence.

This article is an extension of previously-published conference paper \cite{zhou2016reliable} with a new review of the relevant state-of-the-art works, new theoretical developments, and extended experimental results. The main contributions can be summarized as below: 
\begin{itemize}[leftmargin=*]
\item A robust scale estimation and correction method have been proposed and its effectiveness has been tested and verified on the public VO dataset and our self-collected dataset. 
\item A quantitative evaluation of several camera height based absolute scale estimation methods have been given in the experimental part. Advantages and disadvantages of different methods have been analyzed and verified on the synthetic and real dataset. 
\item Finally, this paper provides a general introduction and summary of different kinds of absolute scale estimation methods, which gives a guide for future researchers.
\end{itemize}

\section{Camera height based scale estimation}\label{sec:Camera_height_based_scale_estimation}
Based on two or more frames of a monocular camera, the estimated camera motion and 3D map suffer from a so-called similarity ambiguity. Without loss of generality, we consider two-view geometry here. For two-view geometry, the epi-polar constraint is $\mathbf{x}_{2}^{T}\mathbf{F}\mathbf{x}_{1} = 0$,
where $\mathbf{x}_{1}$, $\mathbf{x}_{2}$ are correspondences in two views and $\mathbf{F}$ is the fundamental matrix. Assuming the camera is calibrated, the above equation can be reformulated by using essential matrix as $\mathbf{\hat{x}}_{2}^{T}\mathbf{E}\mathbf{\hat{x}}_{1} = 0$, in which $\mathbf{\hat{x}} = \mathbf{K}^{-1}\mathbf{x}$ is the bearing vector started from the camera center to its corresponding 3D point. The essential matrix can be expressed as $\mathbf{E} = [\mathbf{t}]_{\times}\mathbf{R}$, which has only five degrees of freedom: the rotation matrix $\mathbf{R}$ has three degrees of freedom while the translation $\mathbf{t}$ has only two degrees of freedom because it is up to an overall scale. The estimated camera translation is up to a global scale which is usually chosen to satisfy $\Vert\mathbf{t}\Vert = 1$ for convenience. Based on the estimated camera motion, the 3D scene is only reconstructed up to a global scale. Assuming that an object's length is $b$ based on the 3D reconstruction, the scale factor is defined as $s = b'/b$, where $b'$ is the ground true length of this object. Absolute scale estimation aims at recovering this coefficient $s$.
	
	Camera height is commonly used for scale estimation. Usually, the camera is fixed on a platform and its height (the distance from camera principle center to the ground plane) is unchanged during a certain amount of time. Assuming the ground surface right in front of the camera is flat, the scale can be recovered according to this height information.
	\vspace{-0.25cm}
	\subsection{Ground plane model}
\begin{figure}[!h] 
		\centering
		\includegraphics[width=0.3\textwidth]{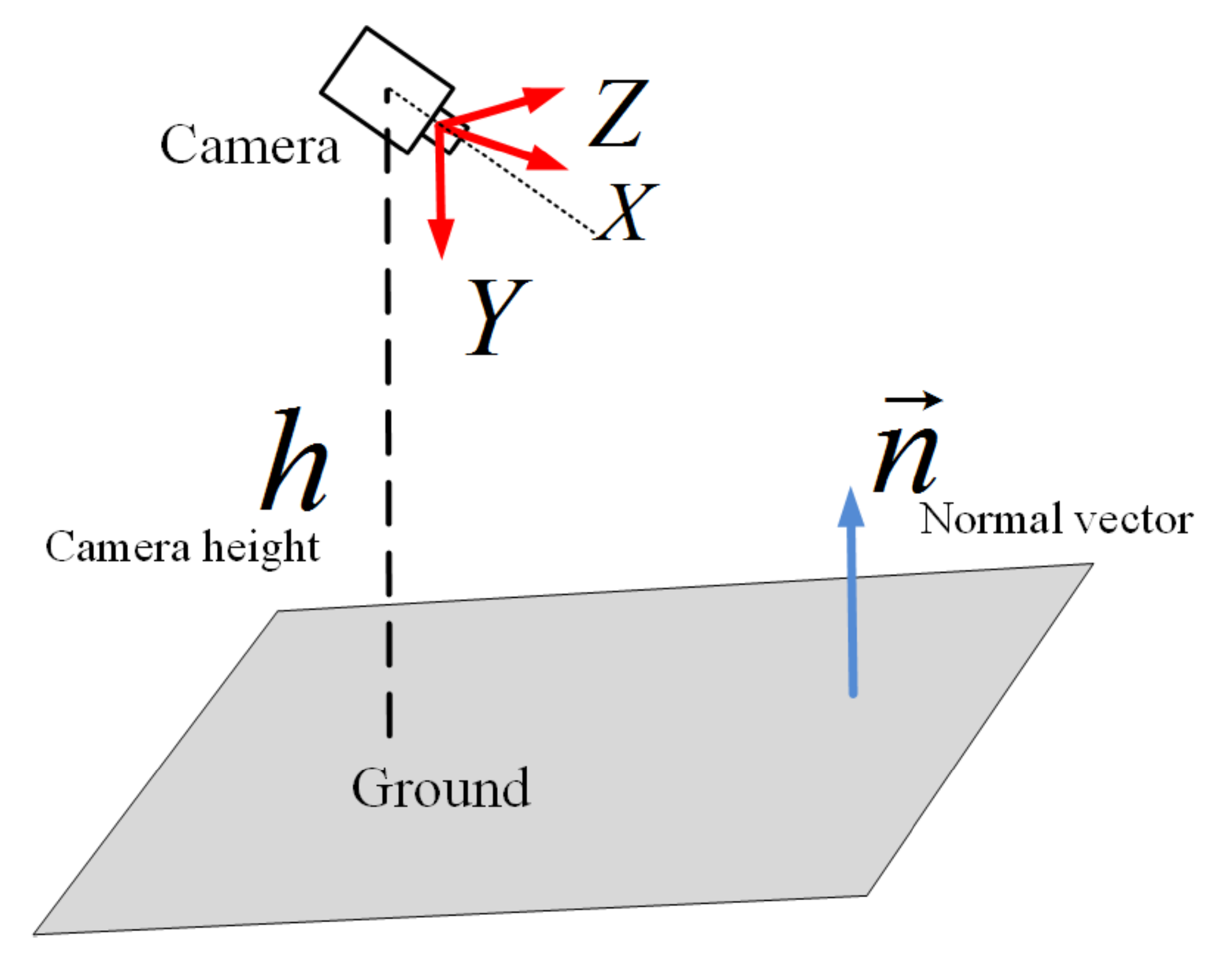}
		\centering
		\caption{Ground plane geometry, where $n$ is the normal of road plane and $h$ is the camera height.}
		\label{fig:Ground_plane}
	\end{figure}

	The camera coordinate at the first frame is assumed to coincide with the world coordinate. Any 3D point $\mathbf{X} = (X,Y,Z)^T$ belonging to ground plane follows the following constraint
	\begin{equation}
	\label{Eq:3D_groundplane_constraint}
	\mathbf{n}^T\mathbf{X} = h,
	\end{equation}
	where $\mathbf{n}$ ($\Vert\mathbf{n}\Vert_{2}=1$) is the normal of ground plane and $h$ is the distance from camera center to ground (as displayed in Fig. \ref{fig:Ground_plane}). Here, we assume that the relative motion $[\mathbf{R}|\mathbf{t}]$ has been estimated (e.g., 8 or 5-points algorithms), in which $\mathbf{R}$ is rotation matrix and $\mathbf{t}$ is the translation up to a scale. The 3D points $\mathbf{X}$ can be triangulated based on 2D features matching and relative motion. The existed camera height based methods can be generally categorized into two groups. 
	\vspace{-0.25cm}
	\subsection{3D plane fitting based scale estimation}
\begin{figure}[!h] 
		\centering
		\includegraphics[width=0.45\textwidth,height=0.25\textwidth]{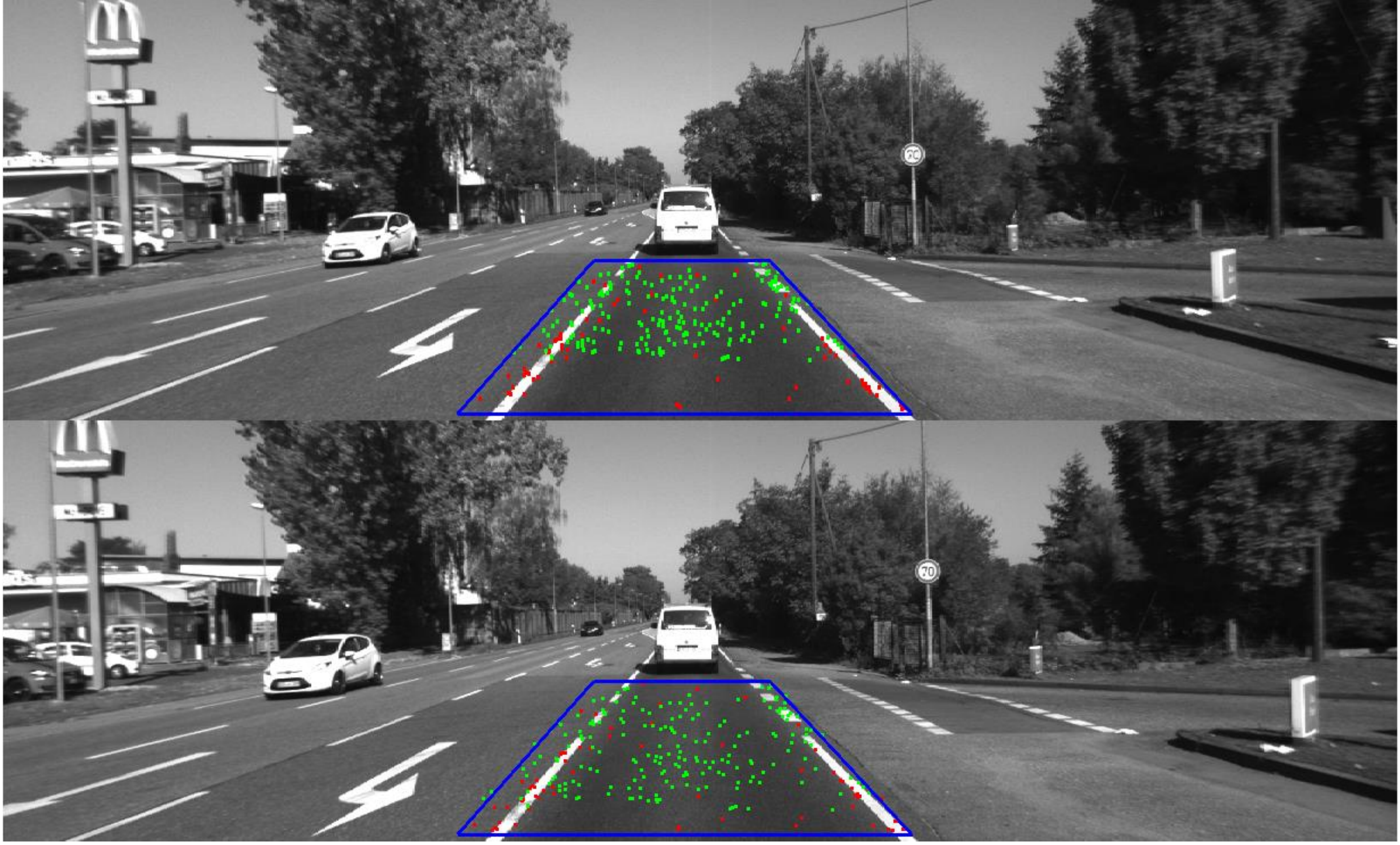}
		\centering
		\caption{Features matching in a pre-defined ROI. The green and red ones represent the inliers and outliers with homography fitting.}
		\label{fig:ROI}
	\end{figure}

	In the camera coordinate, the 3D ground plane (as Eq. \eqref{Eq:3D_groundplane_constraint}) can be fitted from a group of reconstructed 3D points. For robust estimation, a Region of Interest (ROI) as displayed in Fig. \ref{fig:ROI} (blue rectangle) is pre-selected. There are 3 degrees of freedom for ground plane, which comes from the normal direction $\mathbf{n}$ (where $||\mathbf{n}|| = 1$) and the distance $h$. Theoretically, 3 points are the minimum configuration for fitting this plane. 
	
	The 3D points can be reconstructed via two-views triangulation by features (e.g., SIFT or SURF) extraction and tracking. Furthermore, bundle adjustment is also employed to refine the 3D reconstruction. For robust estimation, RANSAC (Random Sample Consensus) technique is used to help against outliers. In \cite{grater2015robust}, two different methods are used to compute the normal vector of ground plane: 1) 3-points based RANSAC together with least-squares optimization; 2) vanishing point estimated from special scene structure is also applied for $\mathbf{n}$ estimation. Then the two normal vectors are fused and tracked by a Kalman Filter for the scale estimation in next frame. 
	
	Another ground plane fitting method was proposed in Libviso2 \cite{geiger2011stereoscan} by assuming the pitch angle is pre-calibrated and unchanged. Then for each 3D point $i$, a camera height $h_i$ can be computed according to Eq. \eqref{Eq:3D_groundplane_constraint} because $\mathbf{n}$ is known via the pitch angle. However, the computed camera height $h_i$ is not same from different points. To determine an optimal camera height, the following strategy is used:
	\begin{itemize}[leftmargin=*]
		\item  First, $h_{ij}$ is defined as the height difference between point $i$ and $j$. 
		\item  Then for each point $i$, a score $q_i$ is computed as 
		\begin{equation} 
		\label{Minimization_n}
		q_i = { \sum\limits_{j=1,j \neq i}^N \text{exp}(-\mu \triangle{h}^{2}_{ij})},
		\end{equation}
		where $\mu = 50$ and $N$ is points number. This score is used to measure the difference between $h_i$ and all other points. 
		\item Finally, the optimal camera height $h$ equals to $h_i$ who has the maximum score $q$.
	\end{itemize}
	\subsection{2D homography based scale estimation}
	To avoid the uncertainty generated from 3D triangulation process, the ground plane geometry can be described by using 2D Homography matrix.

	For any world point belonging to a plane, the projective homography \cite{hartley2003multiple} $\mathbf{H}$ defines the transformation of the image point from the first frame to second frame as: 
	\begin{equation}
	\lambda\mathbf{x}_2 = \mathbf{H}\mathbf{x}_1,
	\label{Eq:Homography_Matrix}
	\end{equation}
	where $\mathbf{x}_1$ and  $\mathbf{x}_2$ are homogeneous image coordinates in the first and second frames respectively. The homography matrix can be represented by using camera motion and plane geometry information as \cite{malis2007deeper}:
	
	\begin{equation}
	\mathbf{H} = \mathbf{K}(\mathbf{R} + \frac{\mathbf{t}\mathbf{n}^T}{h})\mathbf{K}^{-1},
	\label{Eq:Homography_Matrix_RT}
	\end{equation}
	where $(\mathbf{R}, \mathbf{t})$ is the relative camera pose and $\mathbf{K}$ is the camera intrinsic parameter. 
	
	Theoretically, 4 matched pairs are enough to compute $\mathbf{H}$. Usually, robust strategies such as RANSAC is applied to reduce the influence of matching noise. Then, the estimated $\mathbf{H}$ can be refined via a non-linear optimization method by using all the inliers. Assuming that the camera is pre-calibrated, then Euclidean homography matrix $\widehat{\mathbf{H}} = \mathbf{K}^{-1}\mathbf{H}\mathbf{K}$ can be computed via Eq. \eqref{Eq:Homography_Matrix_RT} easily.
	\subsubsection{$\widehat{\mathbf{H}}$ decomposition for scale estimation}
	From Eq. \eqref{Eq:Homography_Matrix_RT}, we can obviously find that $\widehat{\mathbf{H}}$ includes 8 degrees of freedom, where $\mathbf{R}$ and $\mathbf{t}$ have 5 degrees of freedom and the rest 3 degrees of freedom is included in $\mathbf{n}$ and $h$ respectively. Several approaches have been proposed to recover the camera motion $\mathbf{R},\mathbf{t}$ and ground plane $\mathbf{n}, h$ from the Euclidean homography matrix $\widehat{\mathbf{H}}$ directly. In \cite{faugeras1988motion} and \cite{zhang19963d}, two different kinds of numerical approaches have been proposed to obtain $\mathbf{R}, \mathbf{t}$ by decomposing $\widehat{\mathbf{H}}$ via SVD (Singular Value Decomposition). On the contrary, an analytical method is also deduced in \cite{malis2007deeper} to obtain an explicit solution of $(\mathbf{R}|\mathbf{t})$ from $\widehat{\mathbf{H}}$, which can provide the uncertainty propagation from the homography estimation to the final motion results.
	
	
	Though the homography decomposition method is efficient, it is very sensitive to noise due to several reasons: first, the $\mathbf{H}$ is fitted using noisy feature matches from the low-textured road surface; second, too many parameters are required to be computed from $\widehat{\mathbf{H}}$. Both camera motion $\mathbf{R}$, $\mathbf{t}$, and ground plane geometry ($\mathbf{n}, d$) are required to be recovered, which is another challenge for numerical stability. 
	\subsubsection{Optimization based scale estimation}
	\begin{figure}
		\centering
		\includegraphics[width=0.45\textwidth,height=0.32\textwidth]{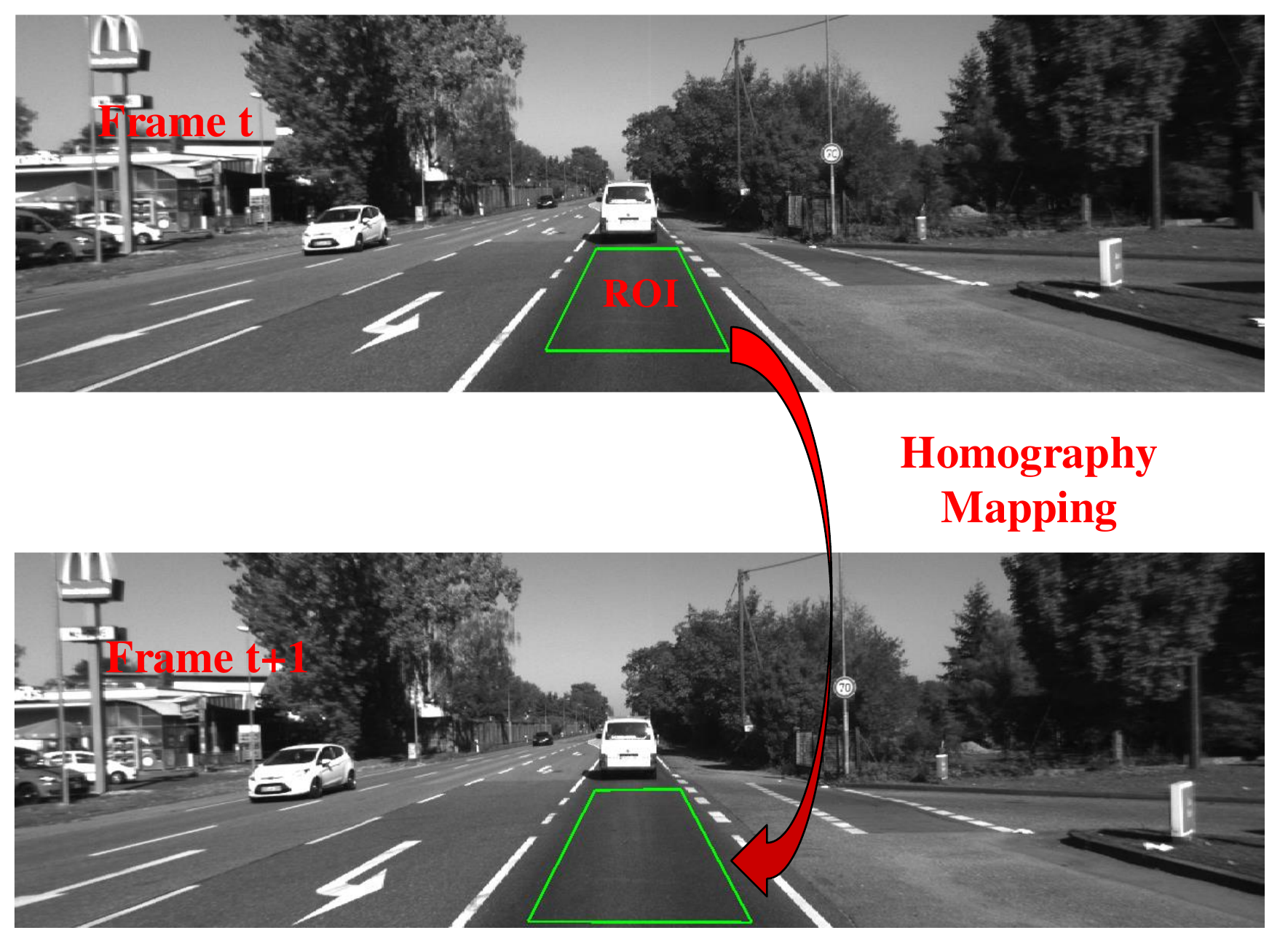} 
		\centering
		\caption{Image mapping via $\mathbf{H}$, which encodes $\mathbf{R,t},\mathbf{n}$ and $d$.}
		\label{Fig:Plane_Guided_Dense_Stereo}
	\end{figure}
	In order to avoid the disadvantages of homography decomposition, many methods proposed to recover camera motion based on optimization \cite{lovegrove2011accurate,kitt2011monocular,song2015high,zhou2016reliable}. Under a plain motion assumption, the energy function between frame 1 and 2 is usually defined on a predefined ROI as 
	\begin{equation}
	\underset{\mathbf{R,t,n},h}{\text{min}} \sum\limits_{i=1}^N \rho(f(\mathbf{x}^{i}_{2})-f(\mathbf{H}_{12}\mathbf{x}^{i}_{1})),
	\label{Eq:Minimization}
	\end{equation}
	in which $\mathbf{x}^{i}_{1}$, $\mathbf{x}^{i}_{2}$ are $i_{th}$ image locations in frame 1 and 2 respectively; $\mathbf{H}_{12}$ is the homography matrix between two frames; $\rho$ represents a certain kind of loss, e.g., $L_1$ or $L_2$ loss; $f(.)$ is a certain function of $\bf{x}$ which can be defined by geometric information (e.g., pixel's location \cite{zhou2016reliable}) or photometric information (e.g., pixel's intensity \cite{song2014robust}). 
	
	Eight parameters are required to be optimized in Eq. \eqref{Eq:Minimization}. Directly solving them together may result in many local minimums due to the non-convexity of the energy function. A simple way of decreasing the possibility of falling into a local minimum is to reduce the number of optimization parameters. Inspired by this idea, many approaches have been proposed to estimate them individually by decoupling the camera motion $\mathbf{R,t}$ and $\mathbf{n}, h$.
	
	A rear camera is used for VO estimation in \cite{lovegrove2011accurate}, where the camera is fixed on the rear of the vehicle and the dominant of the image is road plane. The camera height and pitch angle relative to the ground plane are pre-calibrated. Assuming the camera configuration is unchanged, only the camera motion $\mathbf{R,t}$ is required to be optimized. The loss function is constructed as image intensity error between the wrapped image patch $I_{1}'$ from frame 1 and the real image patch $I_2$ in frame 2. In which, $I_{1}'$ is generated by warping an image patch from $I_1$ via $\mathbf{H}_{12}(.)$ which includes the $\mathbf{R,t}$ and $\mathbf{n}, h$. Finally, the camera motion with absolute scale is obtained by minimizing this objective function. 
	
	\begin{figure}
		\centering
		\includegraphics[width=0.45\textwidth]{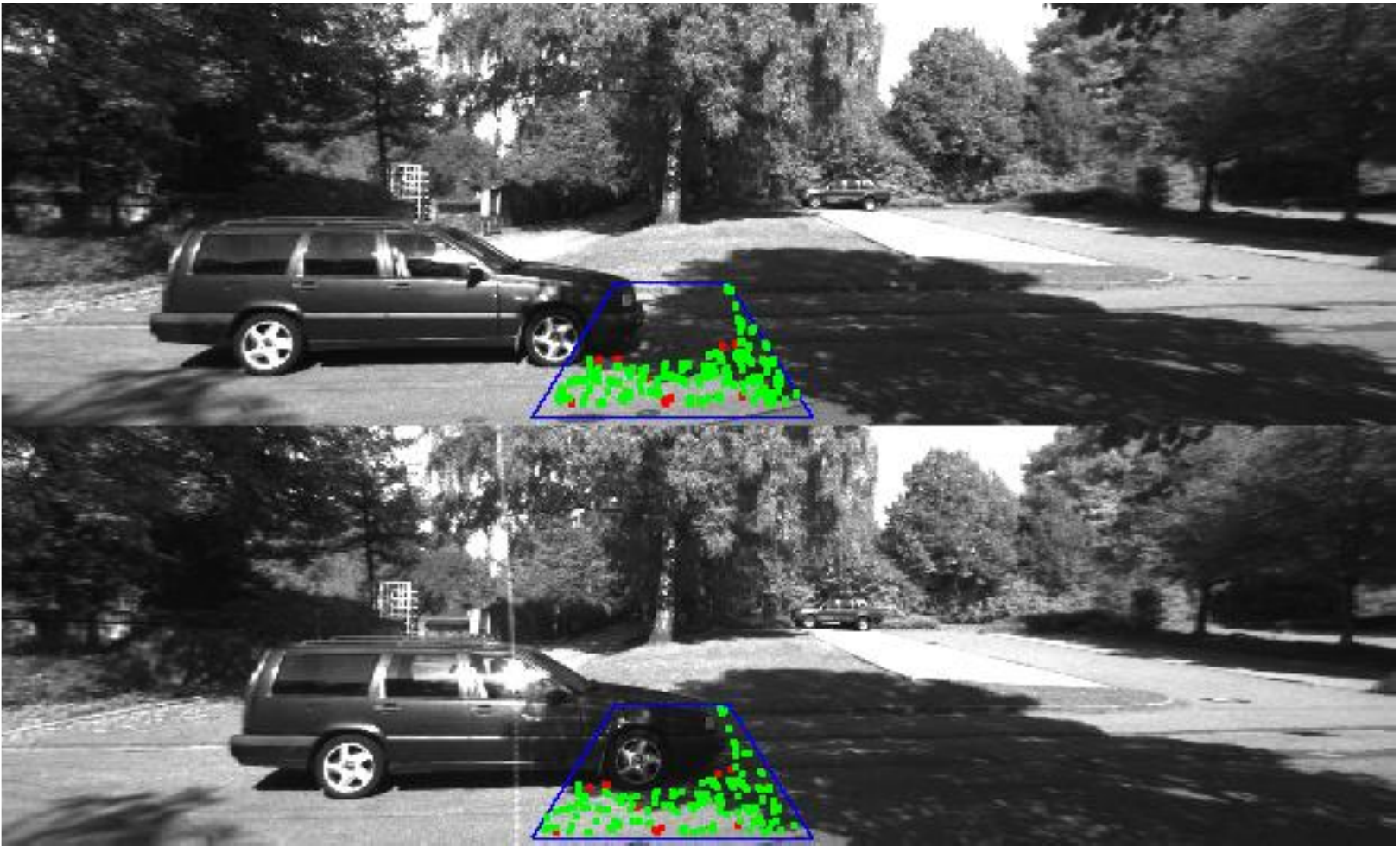} 
		\centering
		\caption{The proposed sparse points based method is robust to noise even some obstacles have been included in the ROI.}
		\label{Fig:Features_Detection_Matching_ROI}
	\end{figure}
	
Alternatively, we can only estimate $\mathbf{n},h$ from Eq. \eqref{Eq:Homography_Matrix} and compute the camera motion $\mathbf{R,t}$ by using other strategies. Song et al \cite{song2015high,song2014robust} proposed to recover $\mathbf{n},h$ by matching a ROI densely between two consecutive frames (as displayed in Fig. (\ref{Fig:Plane_Guided_Dense_Stereo})). Different with the feature matching based method, they match two image patches directly by projecting image patch from frame $1$ to frame $2$. Furthermore, the camera motion is estimated by using other strategies rather than decomposition from $\widehat{\mathbf{H}}$. By doing this, the accuracy and the efficiency of the optimization have been highly improved because the optimization parameters are reduced from eight to three. In addition, other cues are also employed for scale estimation, such as the height of the vehicle, etc.
	\section{Proposed Robust Divide and Conquer Method}\label{sec:Proposed_method}
	The proposed method in \cite{song2015high} has two obvious drawbacks: first, compared with sparse points based matching, the dense matching is inefficient; second, the choice of ROI is very important in this method and it will fail when some non-ground plane objects are included in the ROI as the example shown in Fig. (\ref{Fig:Features_Detection_Matching_ROI}). To get a better ROI, a basic road detector (e.g., \cite{teichmann2018multinet}) can be used to give the prior knowledge of the road first and then we choose the ROI inside these areas. Here, we used the pre-trained model on the KITTI road benchmark \footnote{www.cvlibs.net$/$datasets$/$kitti$/$eva\_road.php} for the road region prediction. In order to handle these drawbacks, we proposed a robust divide and conquer method to recover the scale based on sparse 2D features. Here, divide and conquer represents that the motion parameters (relative pose) in the homography is decomposed from the structure parameters (plane) of the ground plane to improve the stability of the estimation.
	\subsection{Robust scale estimation}
	By doing this, Eq. \eqref{Eq:Minimization} can be rewritten as
	\begin{equation} 
	\label{Eq:Minimization_n}
	\underset{\mathbf{n}}{\text{min}} \sum\limits_{i=1}^N \rho(\mathbf{x}'_i-\mathbf{H}_{12}\mathbf{x}_i)+\rho(\mathbf{x}_i-\mathbf{H}_{21}\mathbf{x}'_i), 
	\end{equation}
	where $\mathbf{x}_i$ and $\mathbf{x}'_i$ are correspondences between two frames. Matrix $\mathbf{H}_{12}$ represents the homography from frame $1$ to $2$, while $\mathbf{H}_{21}$ is the homography matrix from frame $2$ to $1$ and both of them can be obtained from Eq. \eqref{Eq:Homography_Matrix}.
	
	Compared with Eq. \eqref{Eq:Minimization}, only 3 parameters related to the plane geometry are required to optimize in Eq. \eqref{Eq:Minimization_n}. There are several advantages of doing this: \textbf{First}, camera motion estimation is decoupled from plane geometry estimation. Two view epi-polar geometry estimation by using all feature correspondences across the whole image tends to output more reliable and accurate motion estimation than only using feature points on the ground plane. \textbf{Secondly}, the optimization problem is defined on feature correspondences on the ground plane only, which is generally a small patch in the image. Adding $\mathbf{R},\mathbf{t}$ into the optimization will not improve the estimation but deteriorate the estimation. \textbf{Furthermore}, the optimization problem can be solved more efficiently due to the reduced number of variables. 
\textbf{Finally}, fewer parameters will result in less chance to struck at a local minimum.
	
	Experimental result shows that features matching methods proposed in \cite{geiger2011stereoscan} work efficiently and effectively on the texture-less road surface. Epipolar constraint (fundamental matrix is estimated by using features over the whole image) is also employed to remove false correspondences during the features matching process. Furthermore, the correspondences are refined again by estimating a homography matrix $\mathbf{H}_0$ between all the features. Finally, only inliners are taken for the further optimization process. Robust Huber loss is used to define $\rho(.)$ as
	\begin{equation} 
	\rho(r) =
	\begin{cases}
	\frac{1}{2}r^2             & \quad \text{if } r \leq{r_0}\\
	r_0 \left(r-\frac{1}{2}r_0\right)  & \quad \text{otherwise},\\
	\end{cases}
	\end{equation}
	where $r_0$ is a predefined threshold. Nelder-Mead simplex method \cite{lagarias1998convergence} is applied to find the minimum of Eq. \eqref{Eq:Minimization_n}. The initial value of $\mathbf{n}$ and $d$ are computed linearly from $\mathbf{H}_0$ by taking the estimation of $\mathbf{R}$ and $\mathbf{t}$. 
	
	\subsection{Scale refinement}
	Due to the planar road surface assumption is not always satisfied, the filtering technique is always employed to smooth scale estimation results. Kalman filter has been adopted to smooth the ground plane estimation.  
	
	After obtaining the scale in the previous section, Kalman filter is adapted to filter out unreliable scale estimates. The Kalman filter is defined as below:
	\begin{equation}
	\begin{aligned}
	\mathbf{x}_{k} &= \mathbf{F}_{k}\mathbf{x}_{k-1} + \mathbf{w}_{k-1}, &\mathbf{w}_{k} \sim N(0,\mathbf{Q}_{k}), \\
	\mathbf{z}_{k} &= \mathbf{H}_{k}\mathbf{x}_{k-1} + \mathbf{v}_{k-1}, & \mathbf{v}_{k} \sim N(0,\mathbf{P}_{k}), \\
	\end{aligned}
	\end{equation}
	where $\mathbf{x}$ and $\mathbf{z}$ are the state and observation (or measurement) vectors. Matrices $\mathbf{F}$ and $\mathbf{H}$ are the state transition model and observation model. $\mathbf{w}$ and $\mathbf{v}$  are the process noise and observation error which are usually assumed to follow zero mean Gaussian distribution with covariances $\mathbf{Q}_{k}$ and $\mathbf{P}_{k}$. Here the state variable is the ground plane $\mathbf{x} = (\mathbf{n}^T,h)^T$. While $\left\Vert \mathbf{n} \right\Vert = 1$, there are only two free variables of $\mathbf{n}$ to determine. Then the state vector is $\mathbf{z} = (n_x,n_z,h)^T$. In addition, we assume that all the state variables are independent, thus $\mathbf{Q}_{k}$ and  $\mathbf{P}_{k}$ are diagonal matrices.  Here we define state transition model as
	$\mathbf{F} = \left[ \begin{array}{cc}
	\mathbf{R} & \mathbf{t} \\
	\mathbf{0}^T & 1 \\
	\end{array} \right]^{T},$ 
	and the observation model as $ \mathbf{H} = \left[ \begin{array}{cccc}
	1 & 0 & 0 & 0 \\
	0 & 1 & 0 & 0 \\
	0 & 0 & 0 & 1 \\
	\end{array} \right].$
	
	We apply several gating mechanisms to augment the robustness and precision of the proposed algorithm. First, only good initial values are used in the optimization process. An initial value is considered to be good only if the angle between the estimated ground normal $\mathbf{n}$ and prior normal vector is smaller than a certain threshold, such as $5^{\circ}$. Secondly, the estimated value after the optimization will also be discarded if the intersection angle between prior normal and $\mathbf{n}$ exceeds a certain threshold. Only good estimations are used to update the Kalman filter. Our scale estimation approach is summarized in Alg. \ref{alg:scale_estimation}.
\begin{algorithm}[ht!] 
\caption{Scale estimation}
\begin{algorithmic}[1]
\Require{\begin{minipage}[t]{1\textwidth}
- Two consecutive frames $\mathbf{I}_{1}$ and $\mathbf{I}_{2}$;\\
- Camera intrinsic parameters and height;
\end{minipage}}
\Ensure{
~- Estimated scale; 
}
\noindent \begin{raggedright}
\rule[0.15dd]{0.999\linewidth}{1pt}
\par\end{raggedright}
\State $\blacktriangleright\;$Robust sparse features matching between $\mathbf{I}_{1}$ and $\mathbf{I}_{2}$;
\State $\blacktriangleright\;$Robust $\mathbf{R}$,$\mathbf{t}$  estimation with RANSAC;
\State $\blacktriangleright\;$Robust $\mathbf{H}$ estimation in pre-defined ROI;
\State $\blacktriangleright\;$Compute initial $\mathbf{n}$ with Eq. (\ref{Eq:Homography_Matrix_RT}); 
\State $\blacktriangleright\;$Refine  $\mathbf{n}$ by minimizing Eq. (\ref{Eq:Minimization_n});
\State $\blacktriangleright\;$Scale smoothing with Kalman filter.
\end{algorithmic}\label{alg:scale_estimation}
\end{algorithm}
\normalsize

	\section{Scale Correction for VO}\label{sec:Scale_correction}

	Local BA can reduce the scale drift based on multi-frames information, while loop closing relies on the revisiting of the same place. However, both of them can not monitor the scale drift timely and actively choose the proper time to correct the scale drift. Another type of scale correction strategy is to detect the scale drift frame by frame by using the prior scene knowledge and trigger the scale correction procedure timely if the scale drift is serious. In \cite{botterill2013correcting}, \cite{frost2016object}, \cite{song2015high} the prior size of the object are used for reducing the scale drift when they are detected in the scene. Obviously, these methods cannot work if no object has been detected. However, as they mentioned in their papers, the object detection provides a limit contribution for scale correction compared with the methods of using camera height and ground plane. 
	
	\begin{figure}
		\centering
		\includegraphics[width=0.4\textwidth]{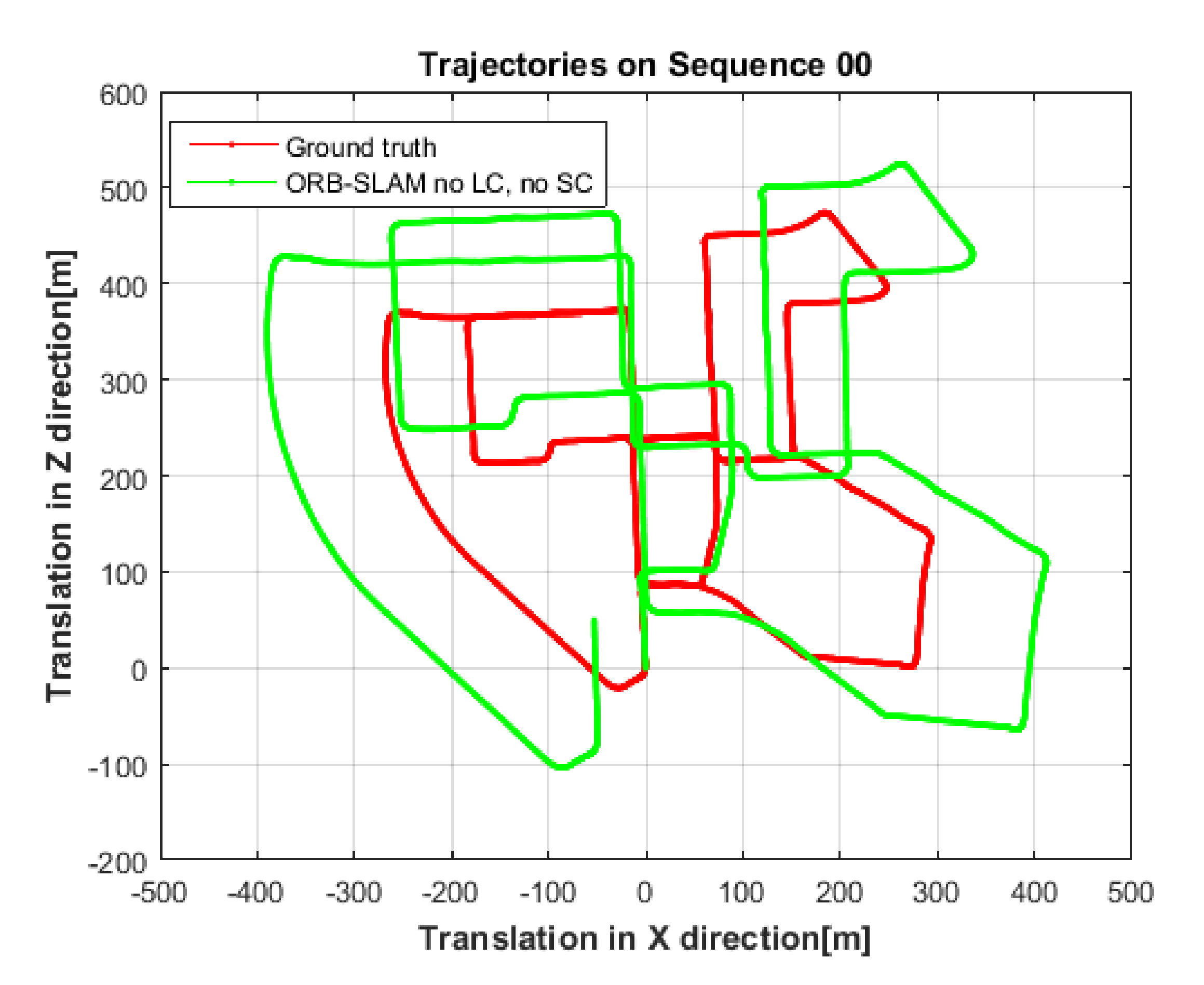} 
		\caption{Camera trajectory estimation results of Monocular ORB-SLAM \cite{mur2015orb} by removing the loop closure on KITTI VO dataset sequence 00. The red curve is the ground truth and green curve displays the estimated result.}
		\label{fig:scale_drift_in_ORBSLAM_00}
	\end{figure}
	
	In \cite{zhou2016reliable}, we propose to use the camera height and ground plane to correct the scale drift for VO/SLAM systems. Furthermore, a robust scale drift detection and correction strategy have also been proposed in this paper. First, the absolute scale is estimated by using the ground plane and camera height for each frame; then a scale drift ratio is computed by comparing the estimated scale and propagated scale in the system; finally, the scale correction procedure is decided to be triggered or not based on this scale drift ratio. 
	
	Although we estimate the absolute scale frame by frame, the scale correction mechanism is only triggered sparingly when the system detects the scale drift ratio over a certain threshold. We choose to correct the scale discontinuously due to several reasons: 1) Per-frame correction is not necessary because the system can hold the scale for a certain period; 2) Per-frame correction will destroy the original VO/SLAM system, such as key-frame selection mechanism; 3) The accuracy of the scale estimation cannot be ensured per-frame as the road surface cannot always be detected reliably.     
	
	\begin{figure}
		\centering
		\includegraphics[width=0.45\textwidth]{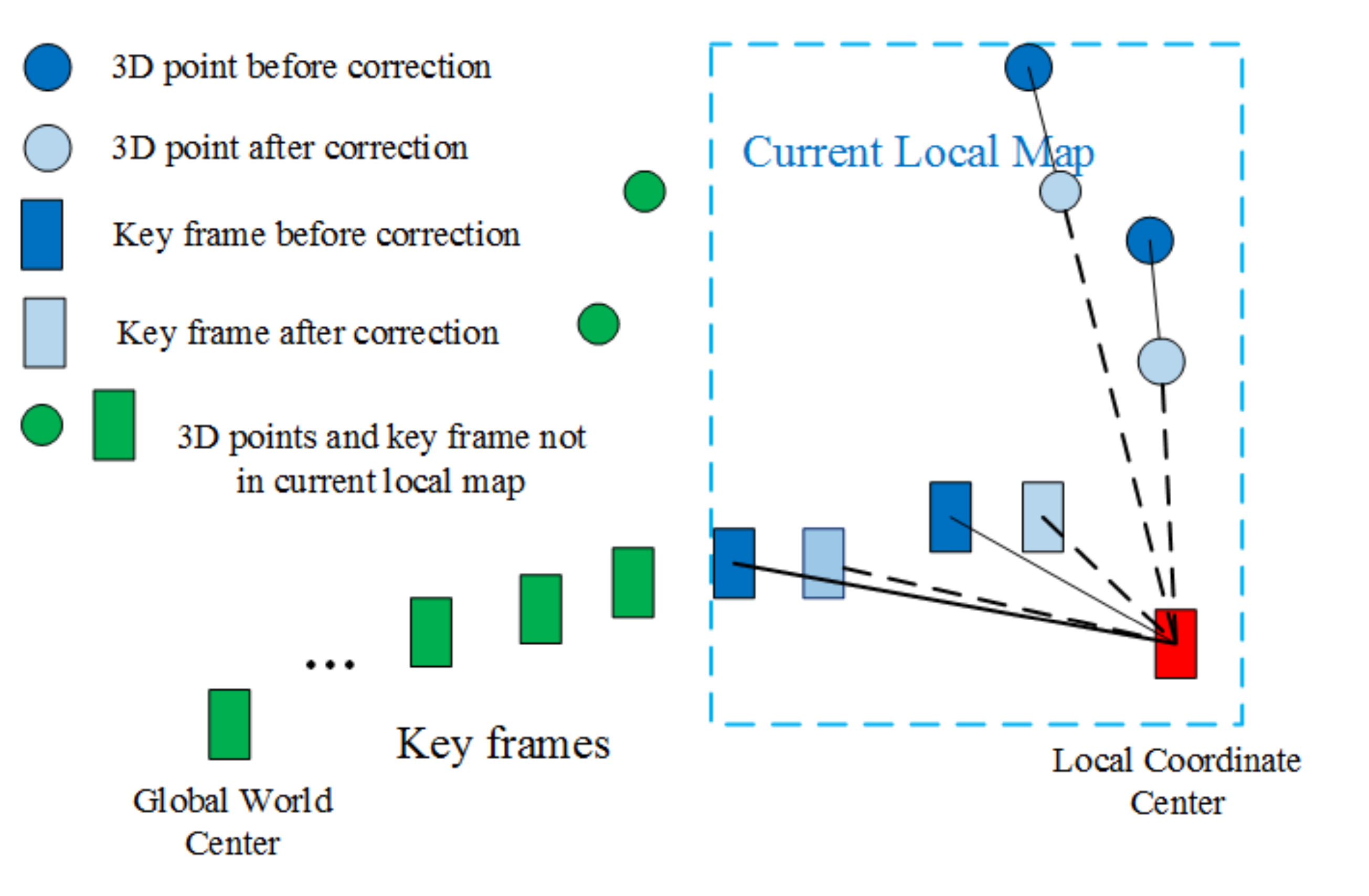} 
		\caption{Sketch of our scale correction strategy. Only the map points and key-frames in the current local map are required to be corrected.}
		\label{fig:ScaleCorrection}
	\end{figure} 
	
	Although the proposed approach can give accurate scale estimation at most of the frames, several criteria are also necessary to make sure that the scale used for correction is accurate enough. We have used the following criteria:   
	1) The estimated ground plane $\mathbf{n}$ should be close to the prior normal direction. 
	2) The velocity should be above a certain threshold. Essential matrix decomposition based camera pose estimation is not accurate under the small motion case.
	3) Only the estimated scale drift ratio satisfies $|\lambda_k-1|>0.075$, scale correction will be triggered.
	
	Once a good scale estimation is ready and the scale correction is required, the correction process will start after the next keyframe is inserted. All the 3D points and key-frames' camera poses in the current local map will be re-scaled. A simple sketch of our correction process is displayed in Fig. \ref{fig:ScaleCorrection}. First, all the 3D map points and the camera poses are transformed into the current local coordinate. Secondly, the local map points and the relative camera poses will be re-scaled by $s$. Thirdly,  the points and camera poses are transformed back into the global world coordinate. Finally, a local bundle adjustment is applied to refine the corrected map points and camera poses. And the updated map points and camera pose are used for the following tracking thread. The main steps of the proposed scale correction strategy are summarized in Alg. \ref{alg:scale_correction}.
	
\begin{algorithm}[ht!]
\caption{Scale correction}
\begin{algorithmic}[1]
\Require{\begin{minipage}[t]{1\textwidth}
- Estimated scale $s$;
\end{minipage}}
\Ensure{
~- Corrected VO; 
}
\noindent \begin{raggedright}
\rule[0.15dd]{0.999\linewidth}{1pt}
\par\end{raggedright}
\State $\blacktriangleright\;$A scale drift ratio $\lambda$ is computed;
\State $\blacktriangleright\;$Scale correction is triggered if $|\lambda_k-1|>0.075$;
\State $\blacktriangleright\;$3D map and pose are corrected after transformed into local coordinate;
\State $\blacktriangleright\;$Local BA is applied after transforming corrected map and pose into global coordinate;
\end{algorithmic} \label{alg:scale_correction}
\end{algorithm}	
\normalsize

	\section{Experimental Results}\label{sec:Experimental_results}
	
	In this section, we will compare different scale estimation methods on the public KITTI VO benchmarking; then the effectiveness of the scale correction is also evaluated on KITTI training VO dataset and our own self-collected fisheye camera dataset. 
	
	\subsection{Scale estimation evaluation}
	In this paper, we mainly focus on comparing three typical ground plane based scale estimation methods. The details of them are described as below:
	\begin{enumerate}[leftmargin=*]
		\item 3D points triangulation based method: scale estimation used in Libviso 2\cite{geiger2011stereoscan} has been chosen for evaluation here because it has been proved to be robust against outliers.
		\item Homography decomposition based method: homography decomposition proposed in \cite{malis2007deeper} is selected for evaluation here.
		\item Optimization based method: dense and sparse matching based method proposed in \cite{song2015high} and \cite{zhou2016reliable} are also used for evaluation here.
	\end{enumerate}
	
	To demonstrate the effectiveness of different kind of scale estimation methods, we evaluated them on a synthetic and real dataset. The KITTI Visual Odometry benchmarking dataset \cite{geiger2013vision} is selected for real experiments evaluation. In addition, in order to make our simulation experiment close to the real scene, the synthetic data is built based on the KITTI image sequences. This KITTI VO benchmark has been divided into training and testing parts respectively. There are 11 sequences in the training dataset, which includes different kinds of traffic scenarios. For the training sequences, the ground truth VO is obtained by fusing high precision IMU and Differential Global Positioning System (DGPS). Based on provided camera trajectory, the ground truth scale is computed by using the relative camera pose between two consecutive frames as $s = \|\mathbf{t}\|_{2}$, where $\mathbf{t}$ is the relative translation vector between two consecutive frames. 
	\subsubsection{Scale estimation on synthetic dataset}
	
	\begin{figure}
		\centering
		\includegraphics[width=0.45\textwidth]{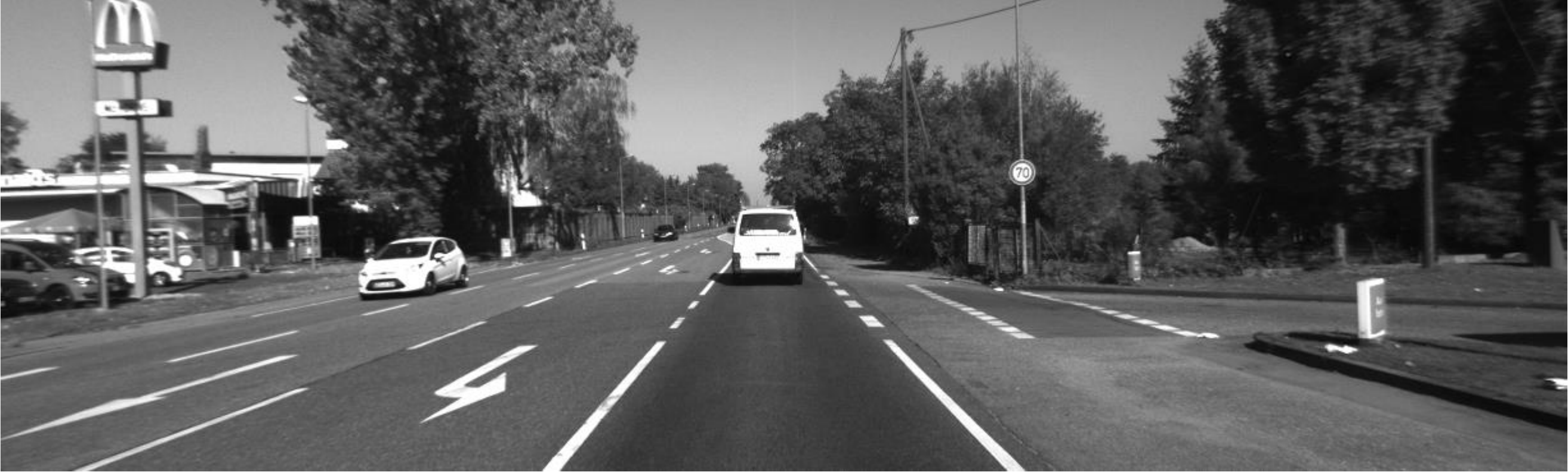} 
		\caption{An example image from sequence 04.}
		\label{fig:An_example_sequence04}
	\end{figure} 
	
	Sequence 04 in the KITTI VO benchmark is selected to generate the synthetic dataset because the road plane is relatively flat in the whole sequence. Sequence 04 includes 271 images and the image frames taken from the left gray camera are taken to generate our synthetic data. Fig.  (\ref{fig:An_example_sequence04}) is an example image of sequence 04. Our synthetic experiment is designed according to the following steps: 
	First, sparse features points are extracted from frame $t$ and on the features points $\mathbf{x}_1$ inside a pre-defined fixed ROI on the ground plane are collected for our following experiments; Then, the ground truth matching points $\mathbf{x}_2$ in the next frame are generated via homography transformation as  $\mathbf{x}_2 = \mathbf{H}\mathbf{x}_1$, where $\mathbf{H}$ is computed by Eq. \eqref{Eq:Homography_Matrix_RT} and
	$\mathbf{R,t}$ use the ground truth relative pose; 
	$\mathbf{n}$ is the normal vector of the road plane assumed to be $[0, 1, 0]^T$; $h$ is set as the real camera height; $\lambda$ is a scale factor to keep $h_{33} =1$. In addition, Gaussian white noise in different levels is added on features $\mathbf{x}_2$ to test the stability of different types of methods. Furthermore, these approaches are also tested with two different speed modes: low and high. At the low-speed mode, the camera moves around \SI{12.5}{\km/\hour}, while \SI{50}{\km/\hour} at the high-speed mode. The speed is controlled by setting the value of $|\mathbf{t}|$. 
	\begin{figure}
		\begin{subfigure}[t]{0.45\textwidth}
			\centering
			\includegraphics[width=0.92\textwidth,height=0.2\textheight]{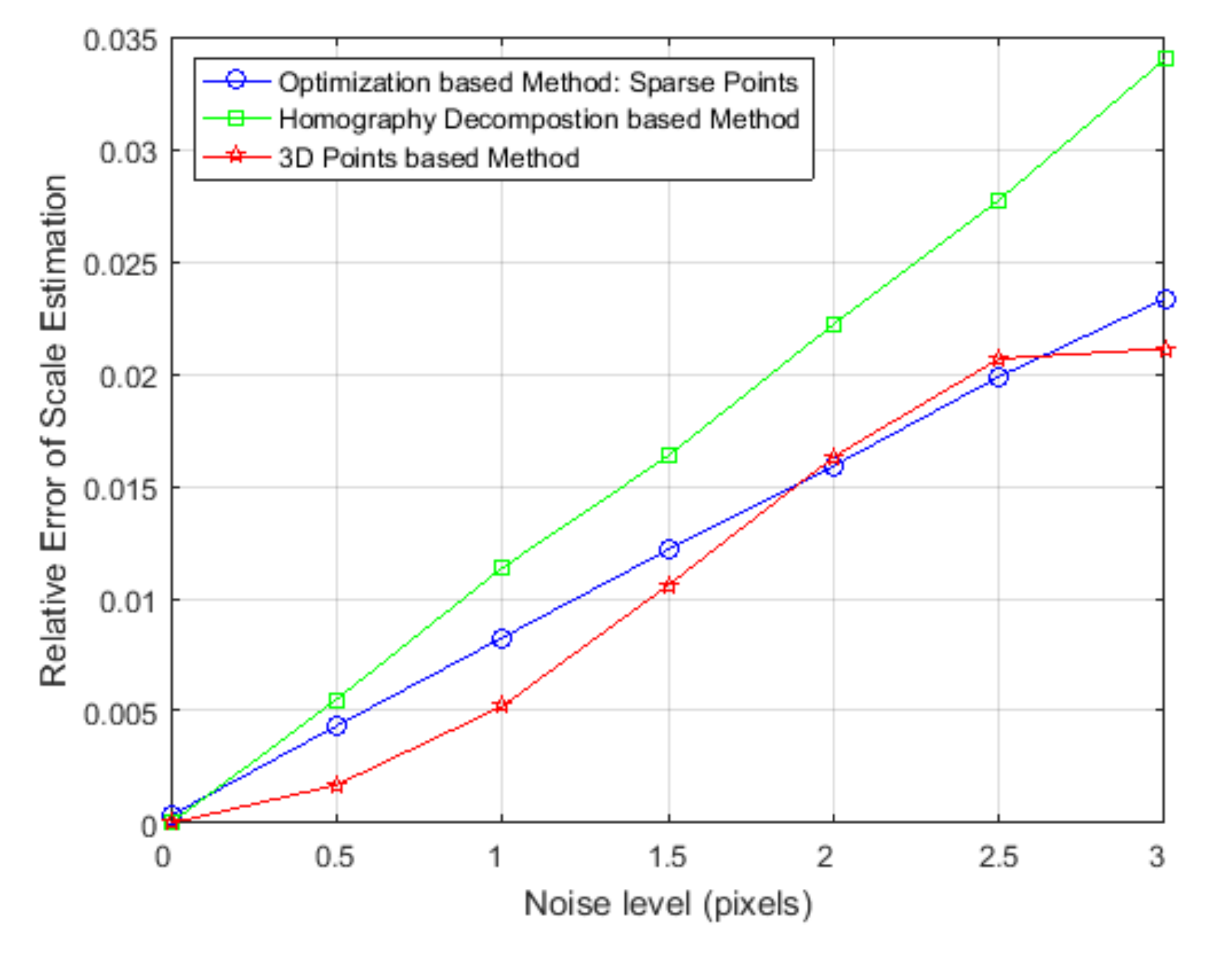} 
			\caption{Scale estimation at high speed (\SI{50}{\km/\hour}).}
			\label{subfig:Scale_on_Synthetic_dataset_high_speed}
		\end{subfigure}\\
		\begin{subfigure}[t]{0.45\textwidth}
			\centering
			\includegraphics[width=0.92\textwidth,height=0.2\textheight]{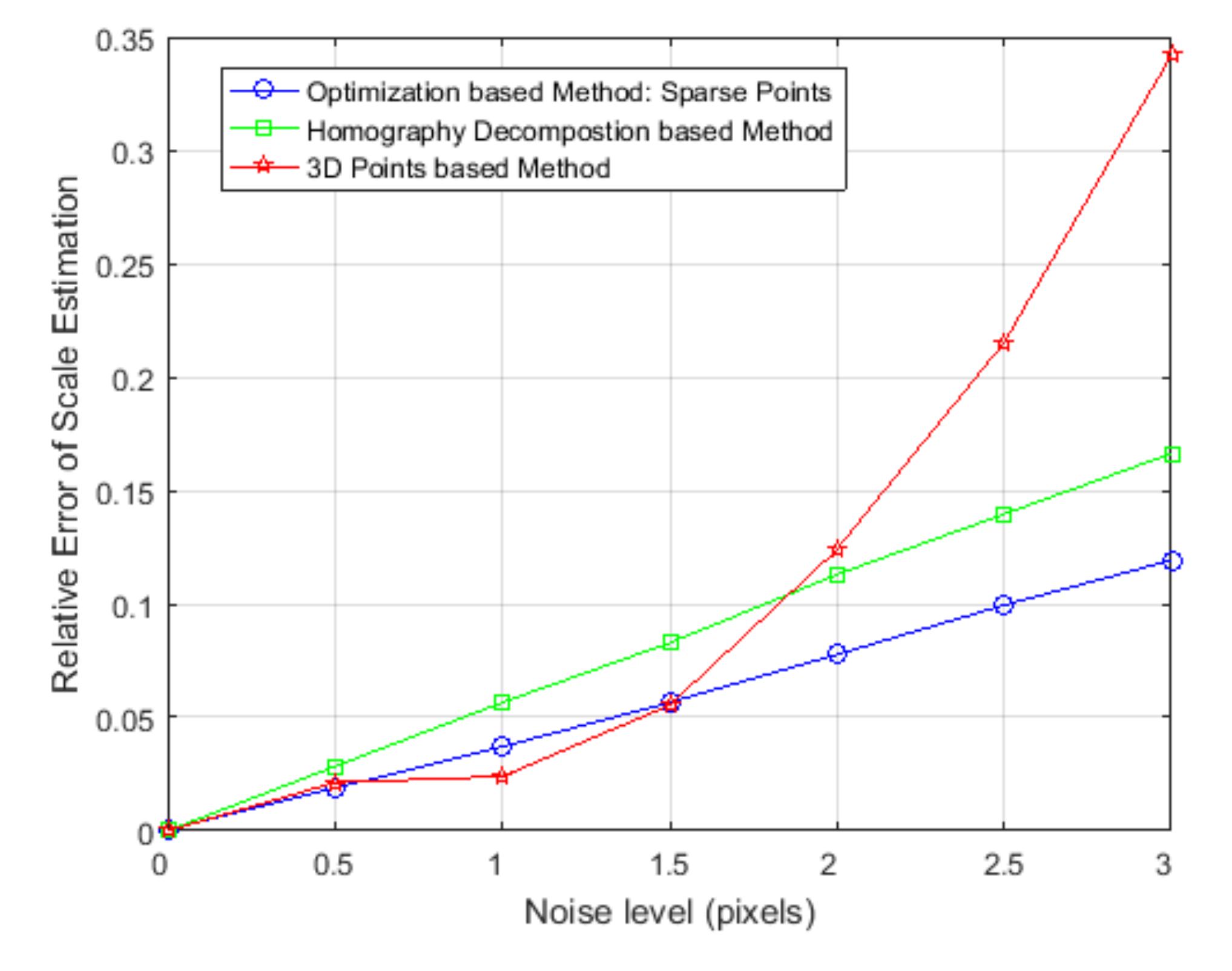} 
			\caption{Scale estimation at low speed (\SI{12.5}{\km/\hour})).}
			\label{subfig:Scale_on_Synthetic_dataset_low_speed}
		\end{subfigure}
		\caption{Scale estimation results on the synthetic experiments. Red line represents the results of the 3D points based method; blue and green lines represents the results of homography decomposition and sparse points optimization based methods respectively.}
		\label{fig:Scale_on_Synthetic_dataset}
	\end{figure}
	\subsubsection*{Quantitative evaluation}
	The scale estimation results are displayed in Fig. (\ref{fig:Scale_on_Synthetic_dataset}), while sub-fig. (\ref{subfig:Scale_on_Synthetic_dataset_high_speed}) and sub-fig. (\ref{subfig:Scale_on_Synthetic_dataset_low_speed}) displays the results at high and low speed modes respectively. In Fig. (\ref{fig:Scale_on_Synthetic_dataset}), red, blue and green lines represent the scale estimation results by using 3D points based method \cite{geiger2011stereoscan}, homography decomposition based method \cite{malis2007deeper} and sparse points optimization based method \cite{zhou2016reliable}. In Fig. (\ref{fig:Scale_on_Synthetic_dataset}), $x$-axis represents the level of the Gaussian noise added on the features points while $x$-axis relative error of estimated scale. 
	
	From these figures, we can easily find that direct homography decomposition based method \cite{klein2007parallel} gives bigger error than sparse point optimization based method \cite{zhou2016reliable} on different noise level and different speed modes. Conversely, 3D points based method \cite{geiger2011stereoscan} performs differently at different speed modes. At the high-speed mode, it gives smallest error among all the three methods, while at the low-speed mode, its error increases dramatically with the increase of the noise's level, especially when the noise is bigger than one pixel. The performance of the 3D points based method can be explained as when the camera moves fast, the 3D triangulation is relatively accurate because the base line between two consecutive frames is large; when the camera moves slow, the 3D triangulation results are inaccurate due to its small baseline. 
	\subsubsection*{Computation time evaluation}
\begin{figure}[ht!]
		\centering
		\includegraphics[width=0.45\textwidth]{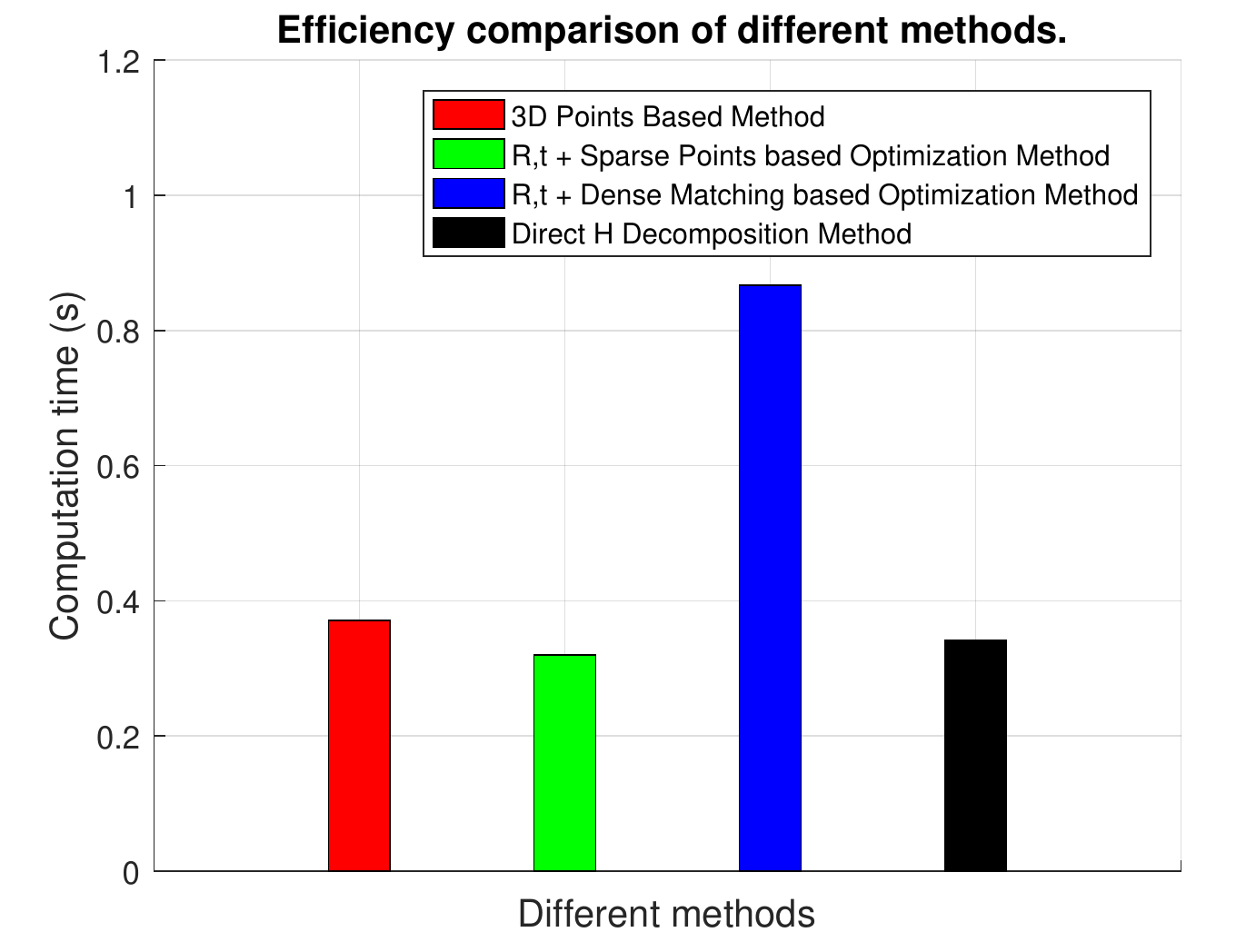} 
		\caption{Computation time of different scale estimation methods.}
		\label{fig:Time_Computation}
	\end{figure}

	The computation time of different approaches is displayed in Fig. (\ref{fig:Time_Computation}). All of them are realized on a standard desktop (Intel Core i7) with Matlab R2016b environment. Fig. (\ref{fig:Time_Computation}) illustrates the average computation time per frame of different scale estimation methods. From this figure, we can see that dense matching method requires the most time compared with other three methods, while the other three methods cost similar time. 
	
	In summary, the sparse point optimization based is relatively robust against the noise on both speed modes compared with the other two methods either the vehicle is at high or low speed. In addition, the sparse point based method cost less processing time compared with the dense matching approach.   
	\subsubsection{Scale estimation on real dataset}
We also evaluate these scale estimation methods on all the real KITTI VO benchmark training sequences. 11 sequences are included in the training dataset, however, only 10 sequences are taken for evaluation here. Sequence 01 is not considered for our evaluation because sequence 01 is taken from a highway with high speed of \SI{90}{\km/\hour} and most of the general feature detection algorithms fail on this sequence.
	
Other 10 image sequences are taken from city suburbs, downtown, and highway, which include about 20000 frames. For each sequence, a fixed size of ROI is chosen for scale estimation. The location of the ROI varies slightly in different sequences depends on the camera installation and the location of the road, however, it is fixed in each sequence. In addition, Kalman filter is also employed to smooth the estimated results. For both the 3D points based method and direct decomposition based method, only the estimated scale has been smoothed by Kalman filter. On the contrary, both the normal vector $\mathbf{n}$ and scale have been taken into the filter and the estimates from the previous frame are used as initial values for the non-linear optimization.  
	
\begin{figure}
		\centering
		\includegraphics[width=0.45\textwidth]{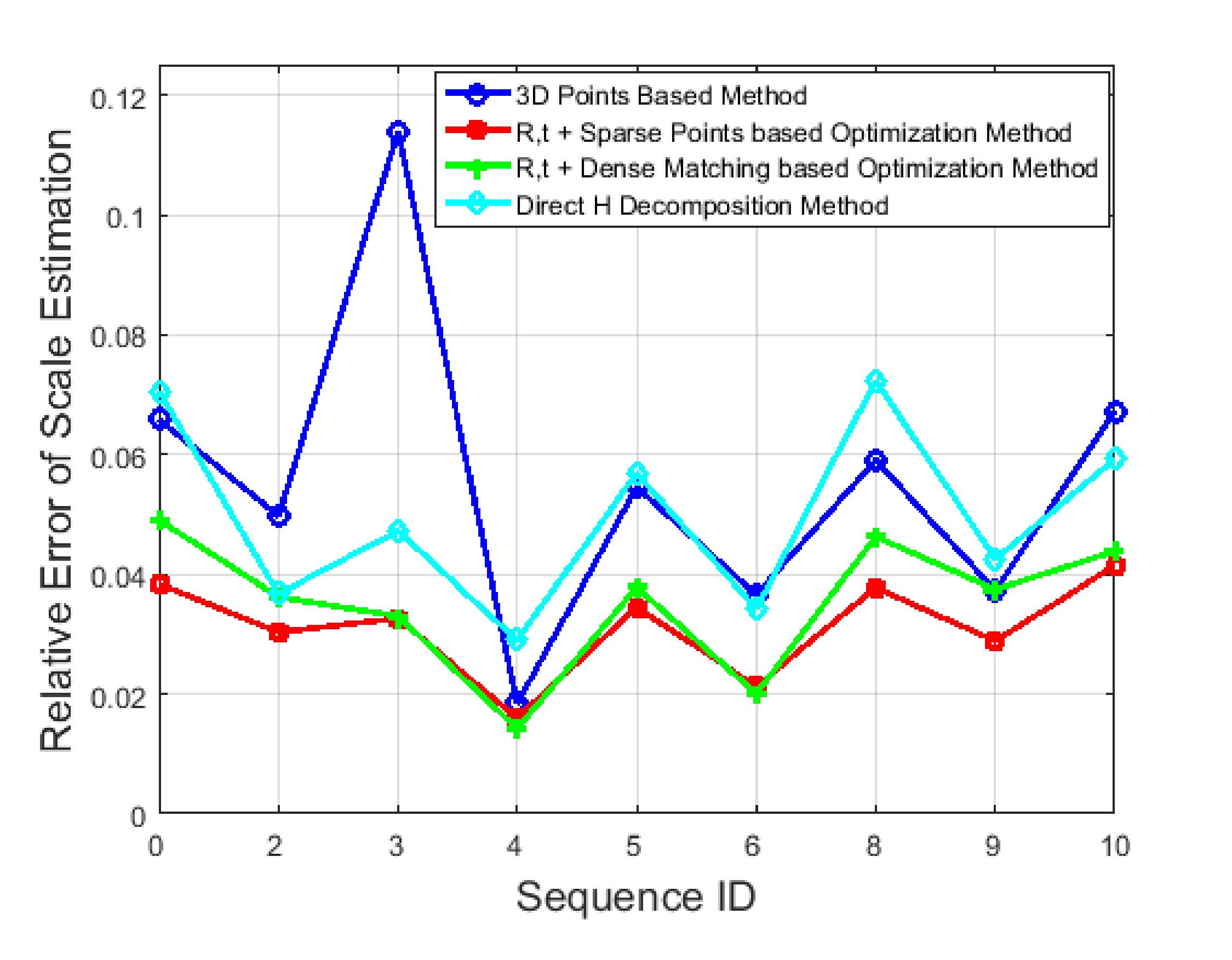} 
		\caption{Scale estimation evaluation on the KITTI VO benchmark training sequences. Blue, greed,  and black lines represent the estimation results by using 3D points based method \cite{geiger2011stereoscan}, sparse \cite{zhou2016reliable} and dense \cite{song2015high} optimization based methods and direct homography decomposition \cite{malis2007deeper} based method.}
		\label{fig:Scale_estimation_evaluation}
	\end{figure}

Fig. \ref{fig:Scale_estimation_evaluation} illustrates the performances of different scale estimation methods on different image sequences, in which Y-axis represents the average scale estimation error and the X-axis gives the sequence ID. The results are drawn with four different colors: blue line represents the results of the 3D points based method; red and green lines represent the results of sparse and dense optimization based methods; cyan line gives the results of the direct homography decomposition based method. From Fig. \ref{fig:Scale_estimation_evaluation}, we can clearly see that sparse points based optimization method gives best results on most of the image sequences, except on sequences 04 and 06, on which dense matching based optimization based method gives slightly better results. Secondly, dense matching based method performances slightly better than 3D points base methods except on the sequence 09. In summary, the sparse and dense optimization based methods give better results compared with 3D points based method and direct homography decomposition based method.
	
	Based on analysis, we can give the following conclusion: first of all, decoupling $\mathbf{R},\mathbf{t}$ and $\mathbf{n}$ can improve the scale estimation results; secondly, 3D points based methods perform relatively well when the baseline is big between two consecutive frame; thirdly, compared with dense matching based optimization method, sparse matching points-based method is more robust when outliers (e.g., other vehicles, curb, sidestep, etc.) are included in ROI.  
	
	\subsection{Monocular VO evaluation on KITTI dataset}
\begin{figure}[ht!]
		\centering
		\begin{subfigure}[t]{0.5\textwidth}
			\centering
			\includegraphics[width=.9\textwidth,height=0.225\textheight]{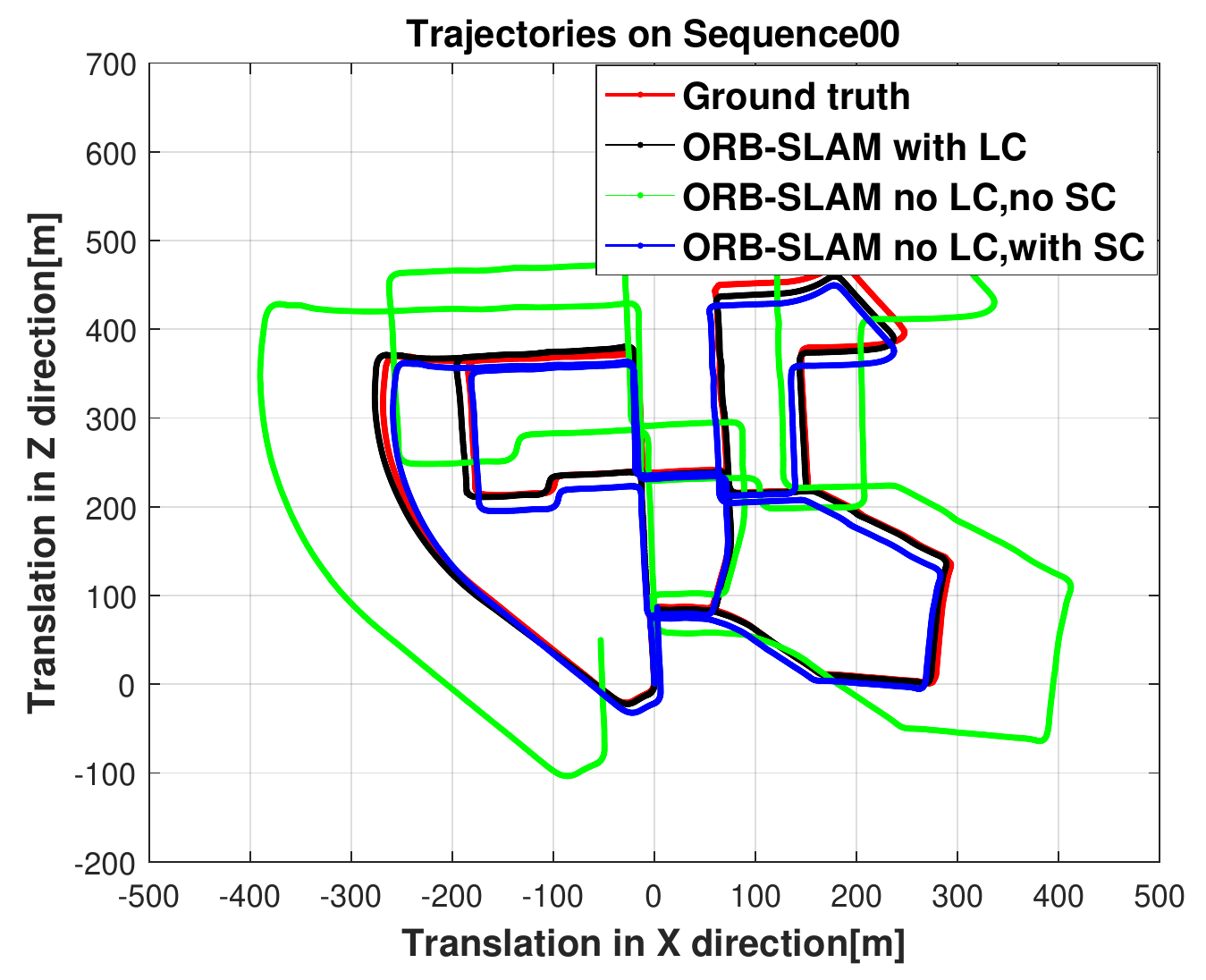}
			\includegraphics[width=.9\textwidth,height=0.225\textheight]{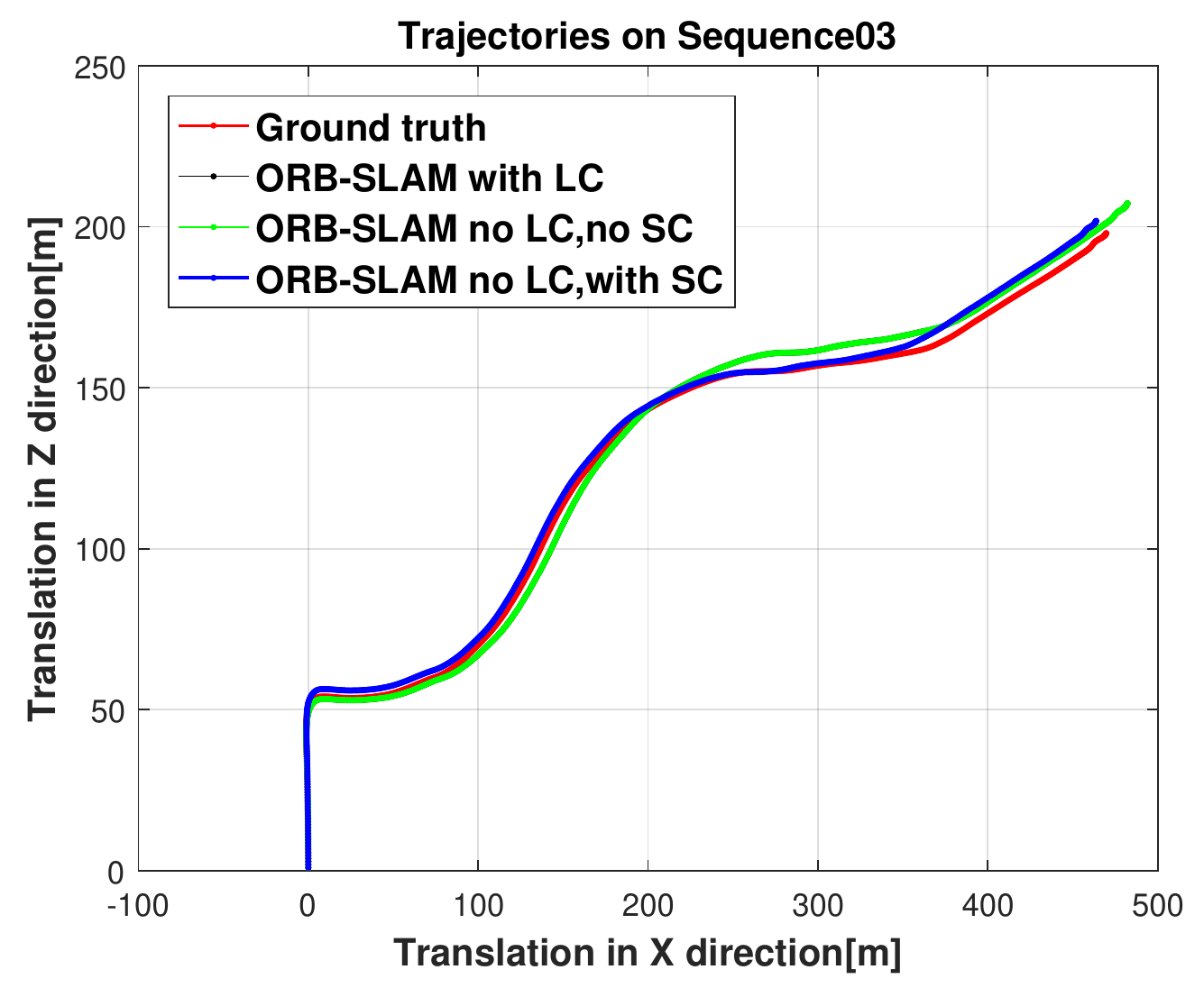}
		\end{subfigure} 
		\caption{Monocular VO comparison with or without scale correction strategy on KITTI dataset sequence 00, 03. Ground truth: the ground truth camera pose; ORB-SLAM with LC \textsuperscript{a}: the original monocular ORB-SLAM with loop closing; ORB-SLAM no LC with SC: monocular ORB-SLAM without loop closing, with our scale correction; ORB-SLAM no LC no SC: monocular ORB-SLAM without loop closing and scale correction.}
		\justifying 
		\small\textsuperscript{a} In the original ORB-SLAM, the absolute scale is not provided. Here, we borrow absolute scale from ground truth pose for system initialization and propagate it in the whole system. 
		\label{fig:Scale_Correction_Evaluation}
	\end{figure}

	In order to test the effectiveness of our scale estimation and scale correction strategy, we test them on the KITTI Visual Odometry benchmark training dataset. Several quantitative experimental results on the training dataset have been displayed on Fig. (\ref{fig:Scale_Correction_Evaluation}). The state-of-the-art ORB-SLAM has achieved surprisingly good performance in the different scenarios with the help of its place recognition and loop closure strategy. However, its performance decreases a lot after removing its loop closure module. Here, we want to use the proposed scale estimation and correction strategy to reduce the scale drift. Fig. (\ref{fig:Scale_Correction_Evaluation}) demonstrates the VO results without or with correction strategy on the two sequences (00, 03). Based on the results, we found that scale drift is serious in the three real traffic sequence, especially in the sequence 00. After scale correction, the estimated camera pose is close to the ground truth trajectory. In addition, the estimated results could be improved if the real loop is detected (e.g., in sequence 00). If there is no loop or the loop is not detection, the ORB-SLAM with or without loop closure will perform the same and the scale drift is serious with the increasing of the cumulative error.  
	
\begin{figure}[ht!]
\centering
\begin{subfigure}[t]{0.5\textwidth}
	\centering
	\includegraphics[width=.95\textwidth,height=0.3\textheight]{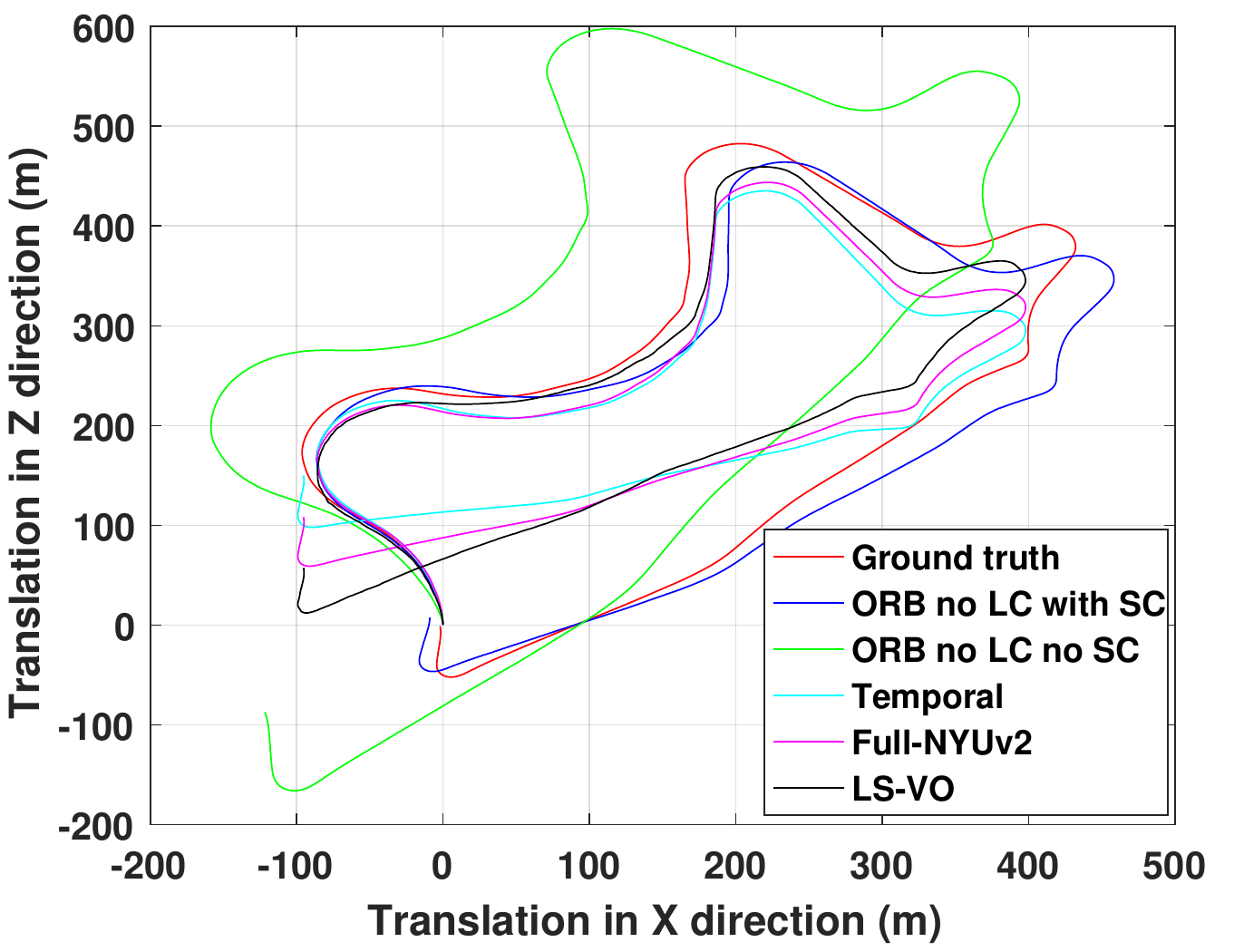}
	\includegraphics[width=.95\textwidth,height=0.3\textheight]{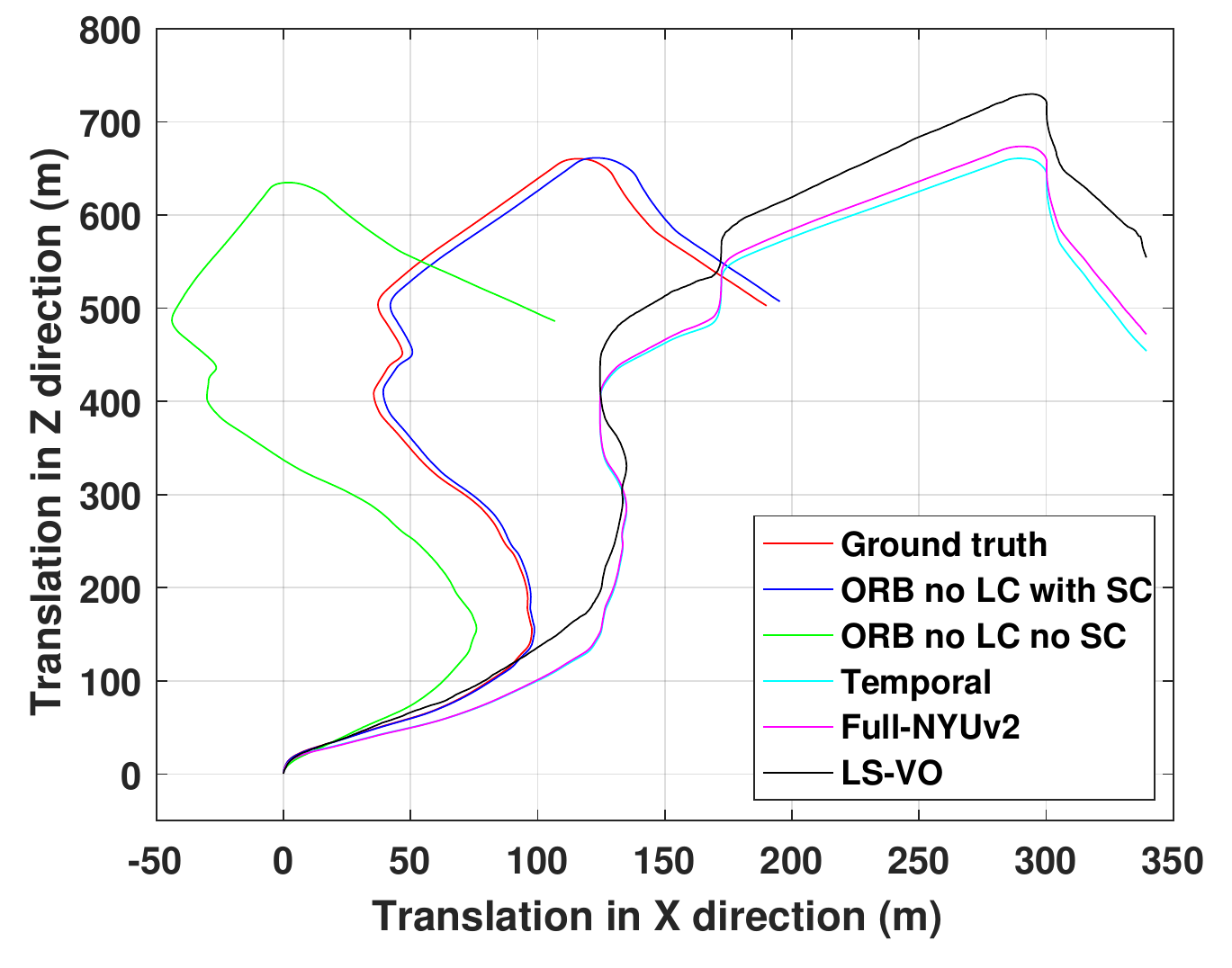}
\end{subfigure} 
\caption{Monocular VO comparison on KITTI data sequence 09, 10. ORB no LC with SC: monocular ORB-SLAM without loop closing, with our scale correction; ORB no LC no SC: monocular ORB-SLAM without loop closing and scale correction; Temporal: model proposed in \cite{zhan2018unsupervised} with additional temporal pairs; Full-NYUv2: model proposed in \cite{zhan2018unsupervised} with additional temporal, stereo, and NYUv2 dataset feature; LS-VO: Latent Space Visual Odometry model proposed in \cite{costante2018ls}.} 
\label{fig:VO_comparison}
\end{figure}

Furthermore, two state-of-the-art deep learning based methods \cite{zhan2018unsupervised,costante2018ls} are also compared here with our method. In \cite{zhan2018unsupervised}, an unsupervised framework has been proposed to learn the depth and VO together. Stereo sequences are used during the training process. It provides both spatial (between left-right pairs) and temporal (forward-backward) constraints. At testing time, only the monocular sequence is used for estimating depth and two-view odometry. In \cite{costante2018ls}, the authors proposed a novel deep network architecture to solve the camera Ego-Motion estimation problem which generally learned features similar to Optical Flow (OF) fields starting from sequences of images. We evaluate the comparison of KITTI sequence 09 and 10, because the rest sequences have been used for training in \cite{zhan2018unsupervised} and \cite{costante2018ls}. Fig. (\ref{fig:VO_comparison}) displays the estimated camera trajectories with different approaches. From the figure, we can find that the proposed method (blue line) is much more closely the ground truth (the red line) compared to other methods. Specifically, the loop closure technique fail to detect the real loop in sequence 09 and 10, therefore the ORB-SLAM with or with LC give the same results.

\begin{figure}[ht!]
		\begin{subfigure}[b]{0.5\textwidth}
			\includegraphics[width=.5\textwidth,height=0.125\textheight]{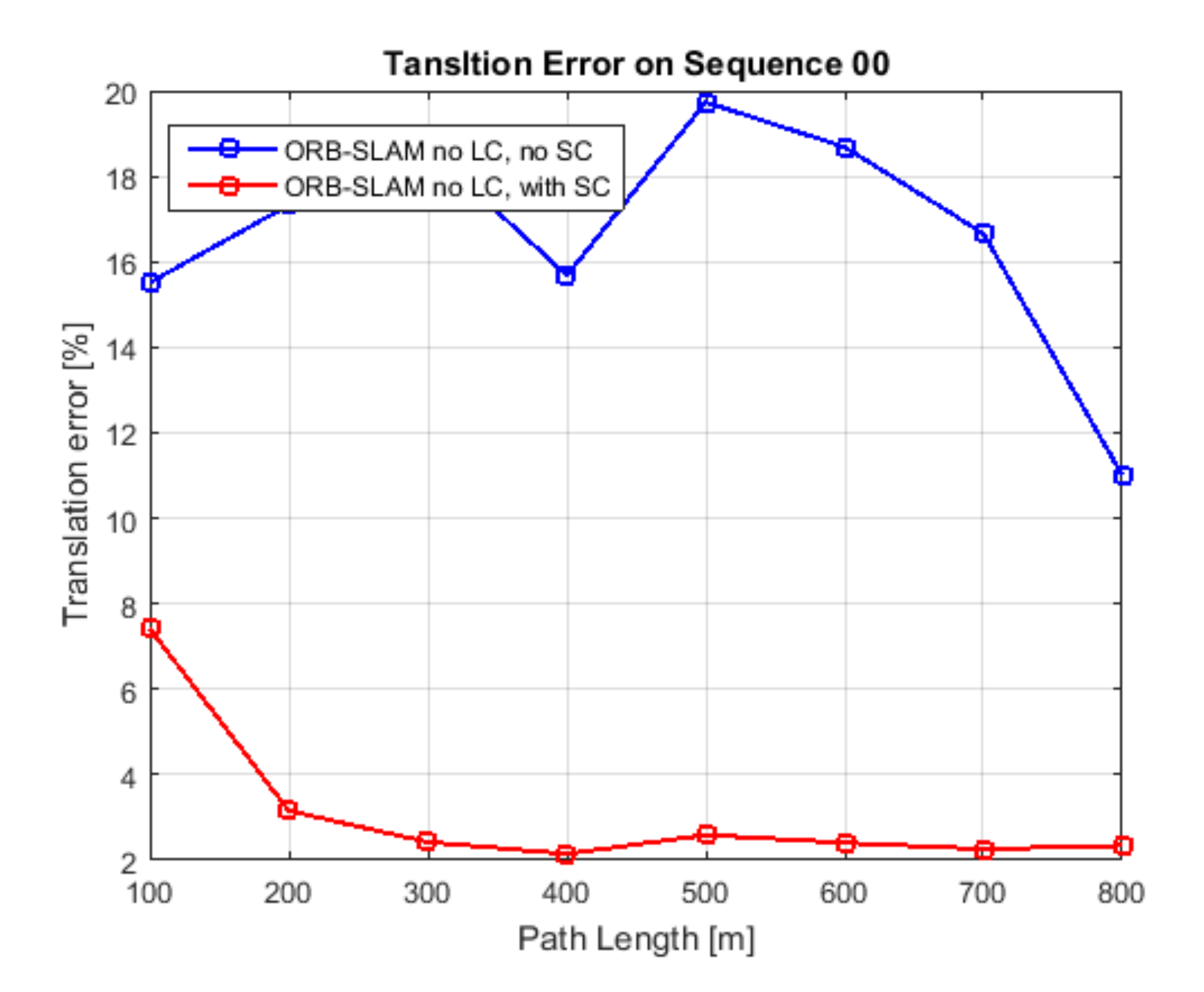}~
			\includegraphics[width=.5\textwidth,height=0.125\textheight]{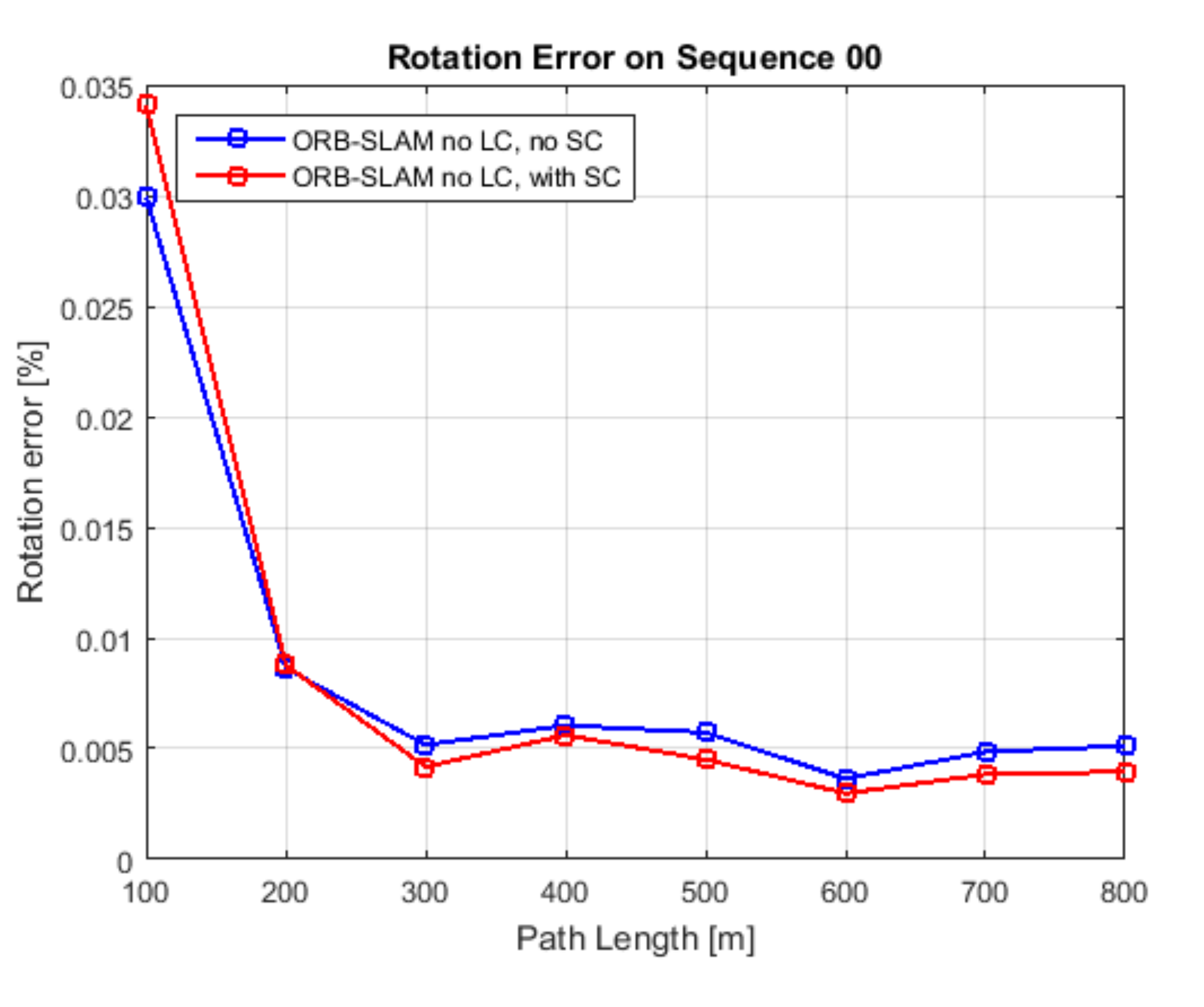}\\
			\includegraphics[width=.5\textwidth,height=0.125\textheight]{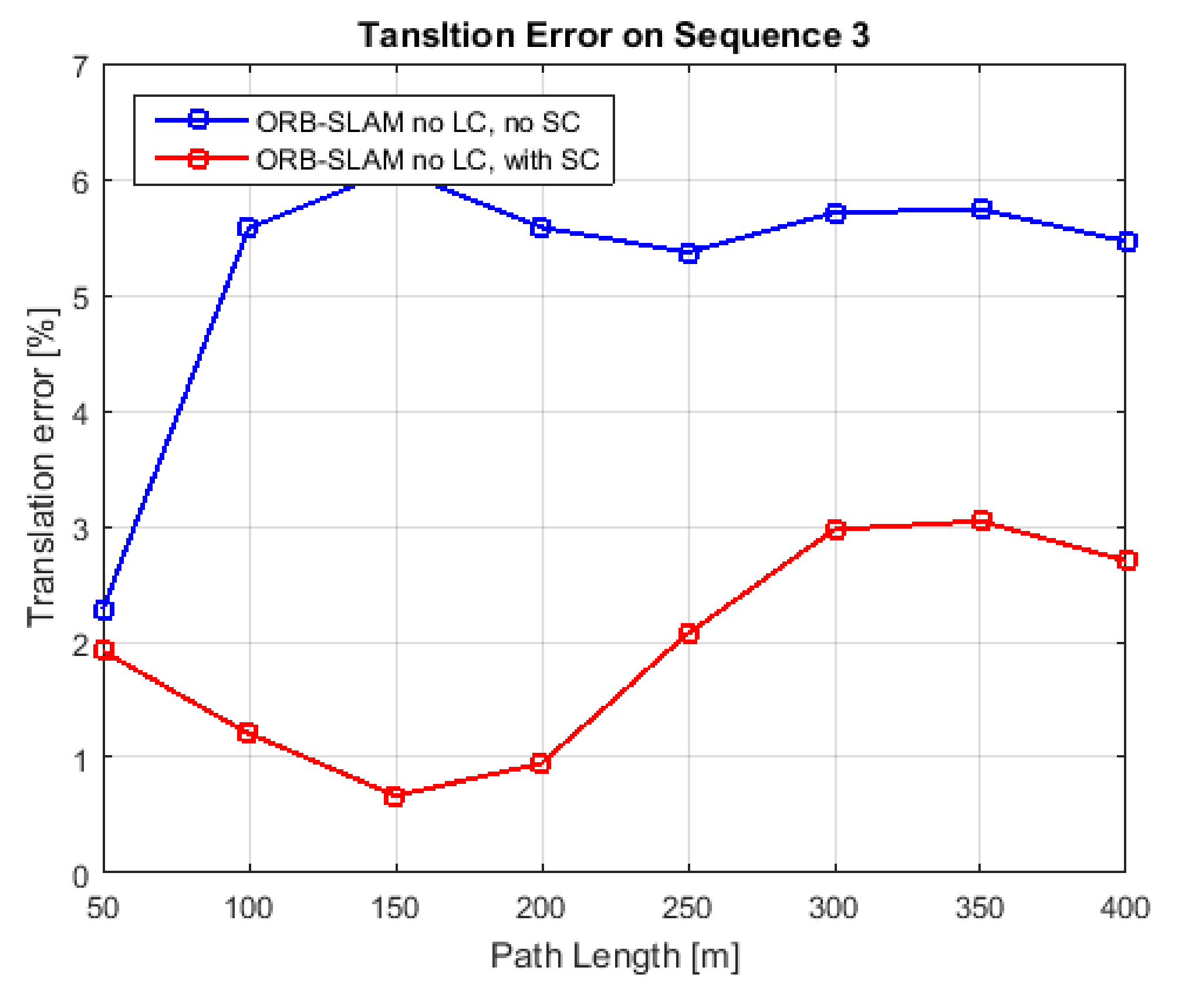}~
			\includegraphics[width=.5\textwidth,height=0.125\textheight]{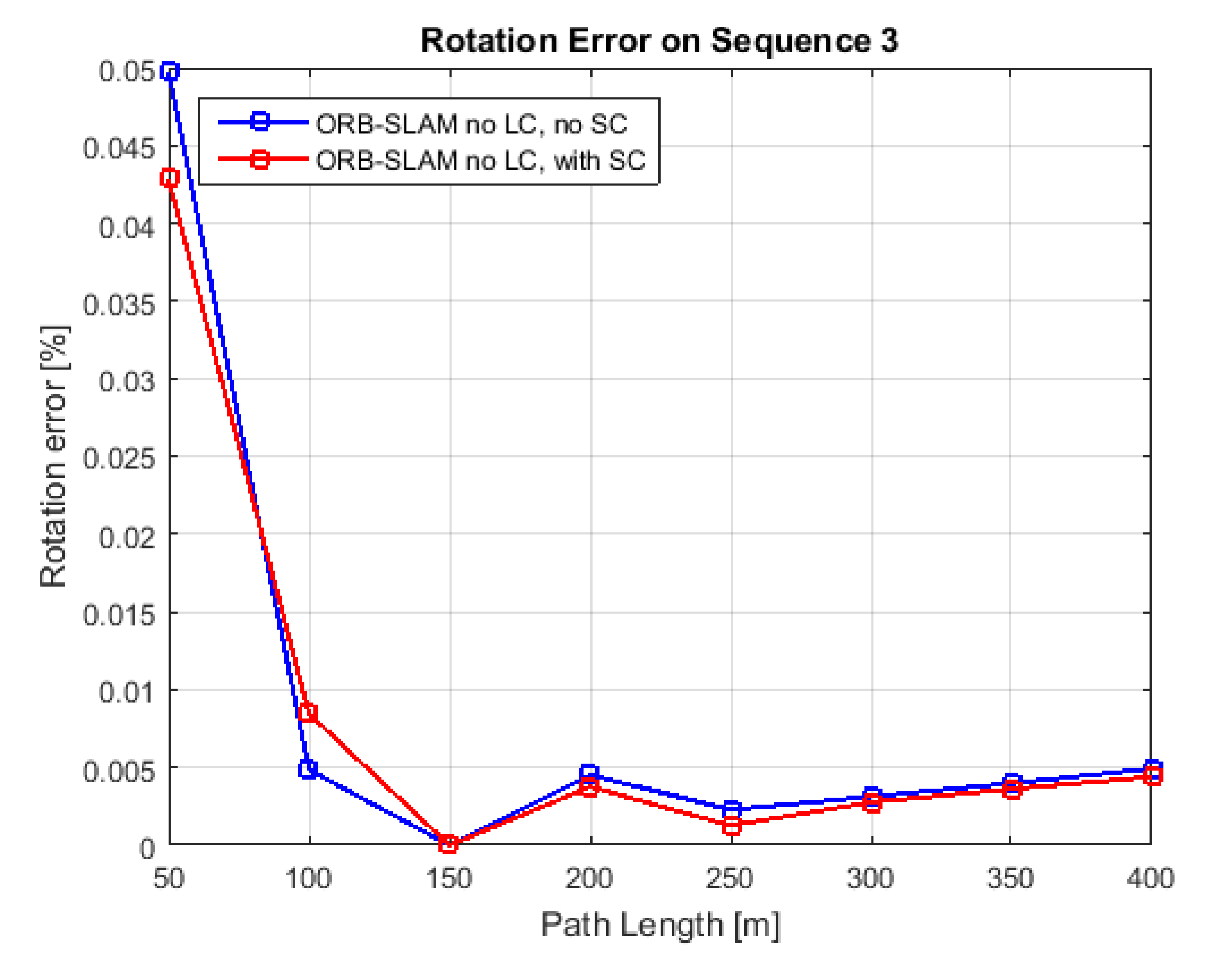}\\
			\includegraphics[width=.5\textwidth,height=0.125\textheight]{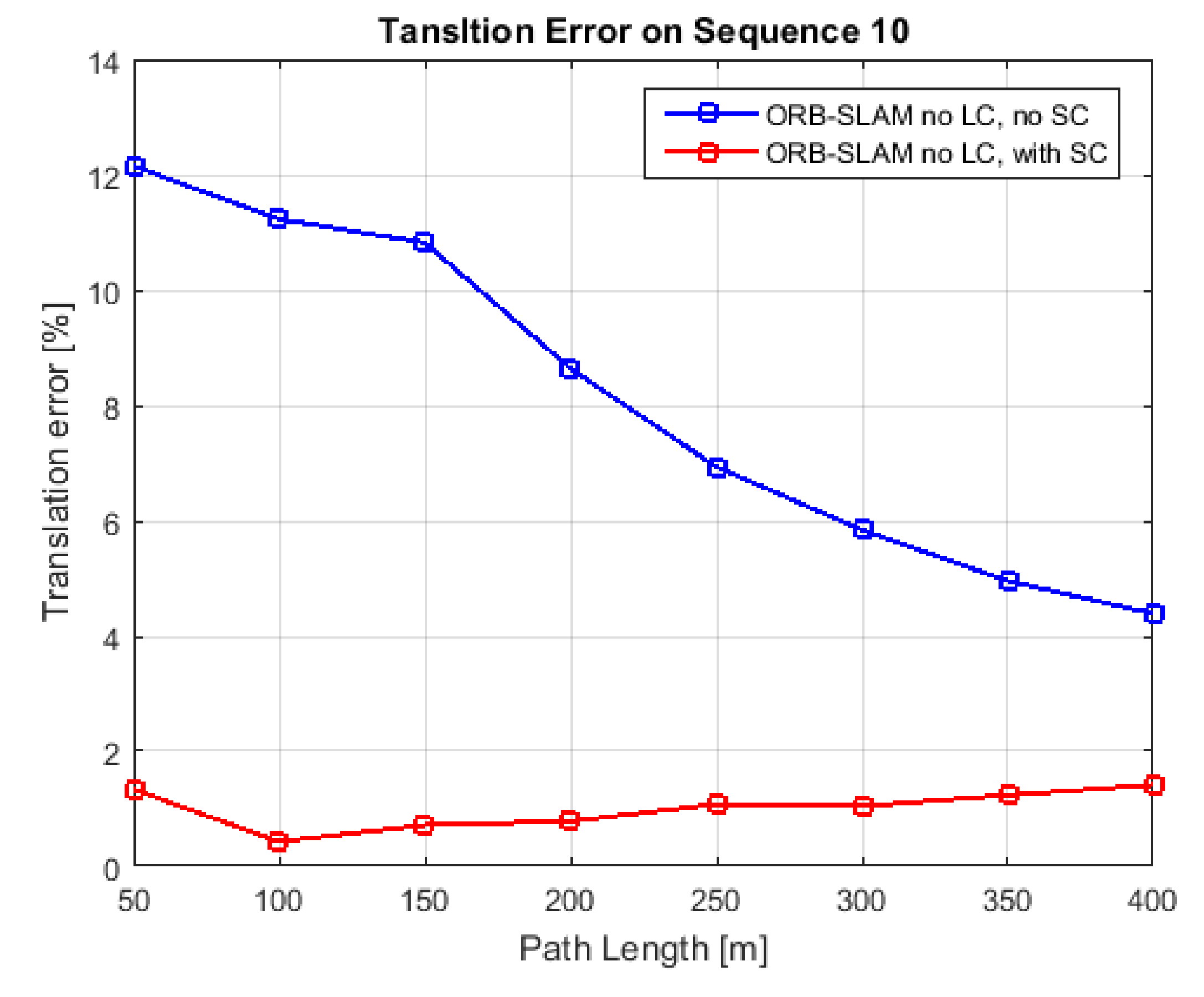}~
			\includegraphics[width=.5\textwidth,height=0.125\textheight]{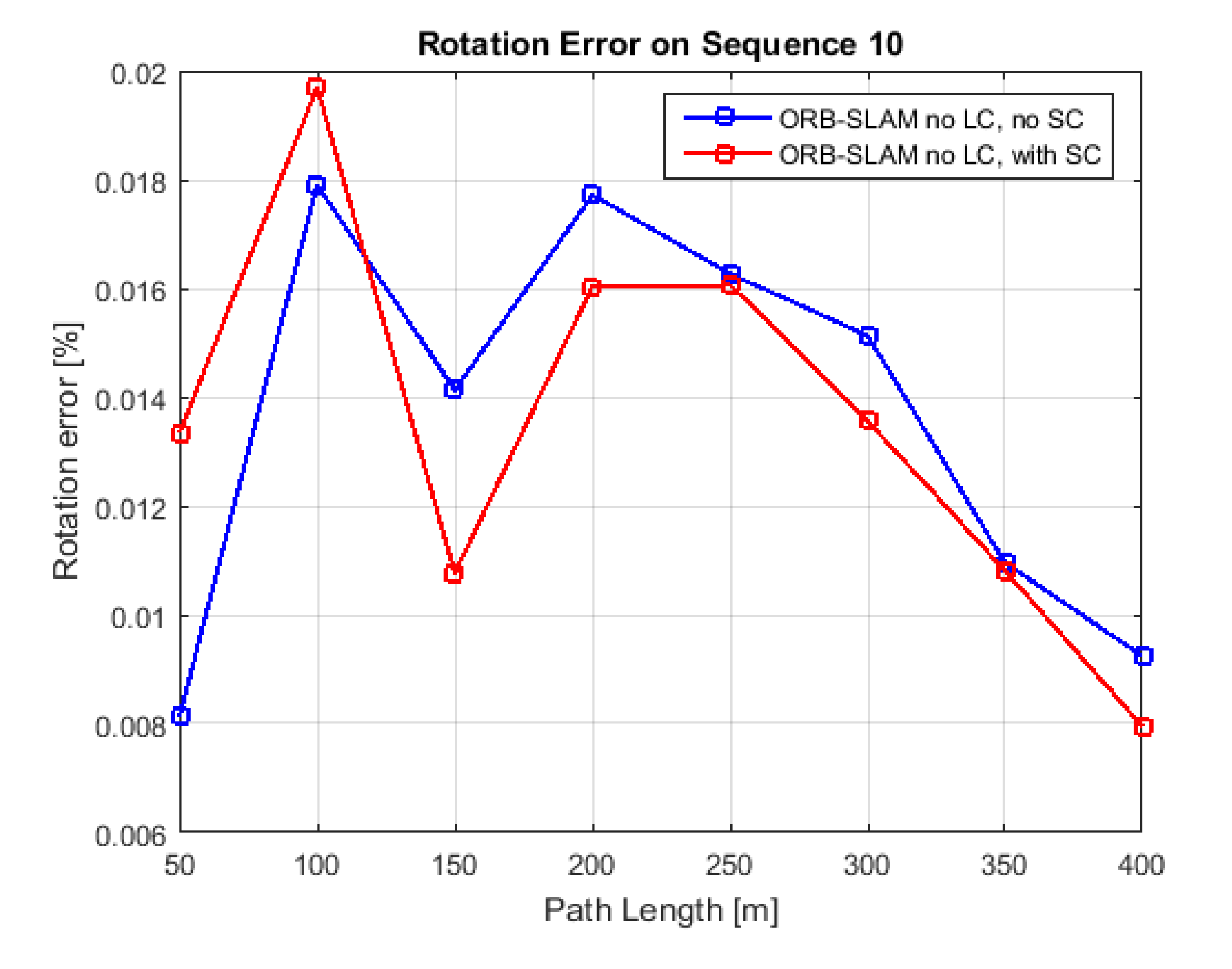}
			\label{fig:Translation_Error}
		\end{subfigure} 
		\caption{Translation and rotation error on sequence 00, 03 10. The errors are computed every 50m for sequence 03, 10 and 100m for sequence 00. Monocular VO comparison with or without scale correction strategy on KITTI dataset sequence 00, 03. ORB-SLAM no LC with SC: monocular ORB-SLAM without loop closing, with our scale correction; ORB-SLAM no LC no SC: monocular ORB-SLAM without loop closing and scale correction.}
		\label{fig:ErrorEvaluation}
	\end{figure}

Quantitative evaluation for VO on the three sequences are shown in Fig. (\ref{fig:ErrorEvaluation}). The red and blue solid lines display the translation and rotation errors with or without scale correction mechanism. From the figure, we can clearly see that our scale correction strategy is extremely effective in reducing the translation error for the three sequences. Especially for the challenging sequence 00 in the urban environment, the translation error has been reduced more than $10\%$ compared with the blue line. The improvement is also obvious in the other two sequences. Because the scale correction is designed to reduce scale drift, it has not many influences for rotation estimation. The rotation errors are kept nearly unchanged with or without scale correction.
	
	\subsection{Visual odometry on self-collected campus dataset}
\begin{figure}
		\centering
		\begin{subfigure}[t]{0.4\textwidth}
			\centering
			\includegraphics[width=0.77\textwidth,height=0.4\textwidth]{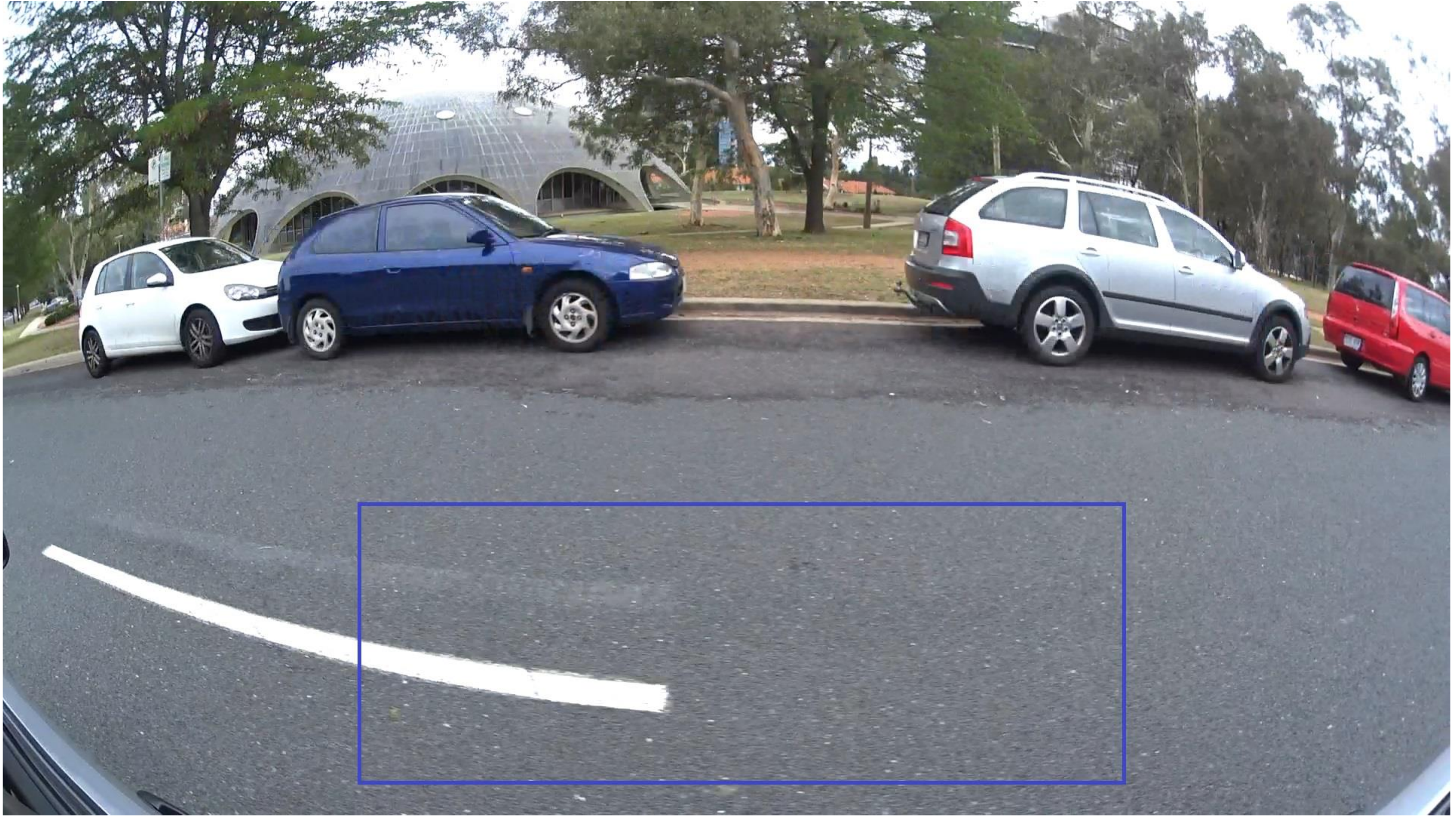}
			\caption{}
			\label{subfig:CampusImage1985}
		\end{subfigure}
		
		\begin{subfigure}[t]{0.4\textwidth}
			\centering
			\includegraphics[width = 0.8\textwidth,,height=0.70\textwidth] {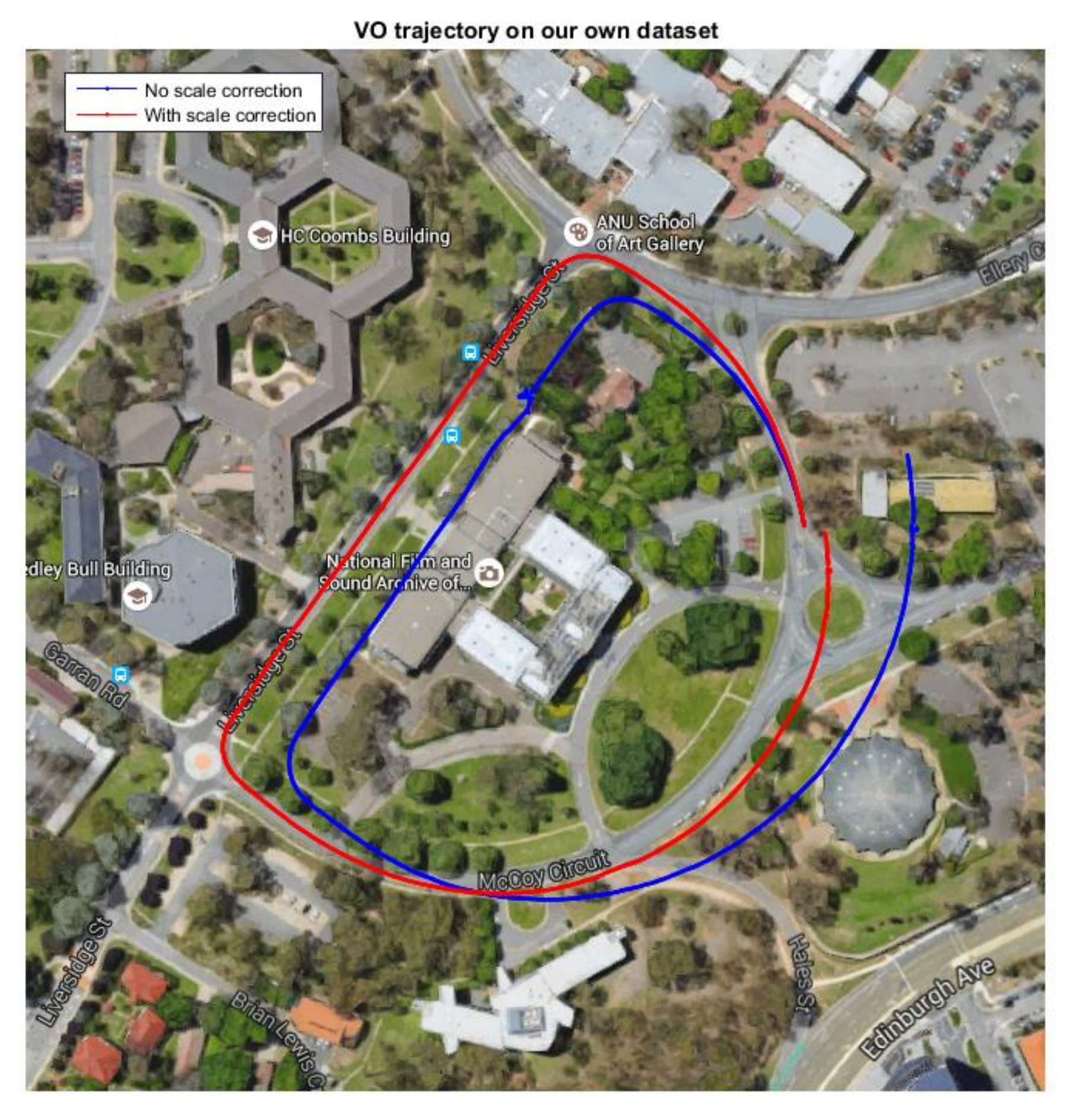} 
			\caption{}
			\label{subfig:Our_UnifySLAM_VisualOdometry}
		\end{subfigure}
		\caption{Scale estimation and correction on our own fisheye image sequence. Red and blue lines denote the VO with and without scale correction strategy.}
		\label{fig:scale_correction_fisheyecamera}
	\end{figure}

	Finally, we test our scale estimation and correction approach on our self-collected campus dataset, which is collected by a monocular fisheye camera (Sony HDR-As200V) mounted on the top of our vehicle. Unlike the KITTI dataset where all the cameras are forward-looking, our camera was installed on the side of our vehicle. The frame rate of our video sequence is 30 fps with the resolution of $1920 \times 1080$ pixels. Similarly, a small fixed ROI is taken at the bottom of the image for our scale estimation as in subfig. (\ref{subfig:CampusImage1985}). The camera is calibrated by using OCamCalib \cite{scaramuzza2006flexible} toolbox offline. And the camera height and pitch angle are also pre-measured. 
	
	As ground truth is not available, camera trajectory is used for qualitative evaluation. In Fig. (\ref{fig:scale_correction_fisheyecamera}), we align the estimated camera trajectory with the Google map, in which the red and blue lines denote the results with or without scale correction. From the figure, we can observe that the estimated VO trajectory by using our scale estimation and correction method is very close to the real road route with a small drift at the end of this sequence. VO without scale correction undergoes serious scale drift in this sequence. The proposed scale estimation and correction strategy are also effective for the fisheye camera. 
	\section{Conclusion and Future Works} \label{Conclusion_and_Future_Works}
	In this paper, we general summary different kinds of camera-based absolute scale estimation methods for monocular visual odometry, which provide a guide for the future researchers. The experimental results on the real dataset show the performance of different approaches. However, all the ground plane based methods will fail when the ground plane is occluded for a long time or road surface doesn't satisfy the plane assumption. The prior scene knowledge-based methods will be invalid if this kind of prior information doesn't exist in some certain environment. In the future, we will focus on estimating reliable absolute scale based on information fusion with two or more other cheap sensors, such as IMU, GPS or lidar, with the camera.
	
	
\section*{Acknowledgment}
This work was supported in part by Natural Science Foundation of China (61871325, 61420106007), ARC grants (DE140100180, CE140100016, DP190102261), Australia ARC Centre of Excellence Program on Robotic Vision. We thank all reviewers for their valuable comments.


	
	\ifCLASSOPTIONcaptionsoff
	\newpage
	\fi
	
	
	
	%
\bibliographystyle{ieeetr}
\bibliography{Reference}

\begin{thebibliography}{10}

\bibitem{mirabdollah2015fast}
M.~H. Mirabdollah and B.~Mertsching, ``Fast techniques for monocular visual
  odometry,'' in {\em German Conference on Pattern Recognition}, pp.~297--307,
  Springer, 2015.

\bibitem{forster2014svo}
C.~Forster, M.~Pizzoli, and D.~Scaramuzza, ``Svo: Fast semi-direct monocular
  visual odometry,'' in {\em IEEE International Conference on Robotics and
  Automation (ICRA)}, pp.~15--22, IEEE, 2014.

\bibitem{mur2015orb}
R.~Mur-Artal, J.~Montiel, and J.~D. Tardos, ``Orb-slam: a versatile and
  accurate monocular slam system,'' {\em IEEE Transactions on Robotics},
  pp.~1147--1163, 2015.

\bibitem{hartley2003multiple}
R.~Hartley and A.~Zisserman, {\em Multiple view geometry in computer vision}.
\newblock Cambridge university press, 2003.

\bibitem{hu2004real}
Z.~Hu and K.~Uchimura, ``Real-time data fusion on tracking camera pose for
  direct visual guidance,'' in {\em IEEE Intelligent Vehicles Symposium (IV)},
  pp.~842--847, IEEE, 2004.

\bibitem{shepard2014high}
D.~P. Shepard and T.~E. Humphreys, ``High-precision globally-referenced
  position and attitude via a fusion of visual slam, carrier-phase-based gps,
  and inertial measurements,'' in {\em Position, Location and Navigation
  Symposium-PLANS}, pp.~1309--1328, IEEE, 2014.

\bibitem{zhang2014robust}
J.~Zhang, S.~Singh, and G.~Kantor, ``Robust monocular visual odometry for a
  ground vehicle in undulating terrain,'' in {\em Field and Service Robotics},
  pp.~311--326, Springer, 2014.

\bibitem{civera2008inverse}
J.~Civera, A.~J. Davison, and J.~M. Montiel, ``Inverse depth parametrization
  for monocular slam,'' {\em IEEE transactions on robotics}, vol.~24, no.~5,
  pp.~932--945, 2008.

\bibitem{desai2016visual}
A.~Desai and D.-J. Lee, ``Visual odometry drift reduction using syba descriptor
  and feature transformation,'' {\em IEEE Transactions on Intelligent
  Transportation Systems}, vol.~17, no.~7, pp.~1839--1851, 2016.

\bibitem{badino2013visual}
H.~Badino, A.~Yamamoto, and T.~Kanade, ``Visual odometry by multi-frame feature
  integration,'' in {\em Proceedings of the IEEE International Conference on
  Computer Vision Workshops}, pp.~222--229, 2013.

\bibitem{scaramuzza2011visual}
D.~SCARAMUZZA and F.~FRAUNDORFER, ``Visual odometry: Part i: The first 30 years
  and fundamentals,'' {\em IEEE robotics \& automation magazine}, vol.~18,
  no.~4, pp.~80--92, 2011.

\bibitem{fraundorfer2012visual}
F.~Fraundorfer and D.~Scaramuzza, ``Visual odometry: Part ii: Matching,
  robustness, optimization, and applications,'' {\em IEEE Robotics \&
  Automation Magazine}, vol.~19, no.~2, pp.~78--90, 2012.

\bibitem{ozyecsil2017survey}
O.~{\"O}zye{\c{s}}il, V.~Voroninski, R.~Basri, and A.~Singer, ``A survey of
  structure from motion*.,'' {\em Acta Numerica}, vol.~26, pp.~305--364, 2017.

\bibitem{fuentes2015visual}
J.~Fuentes-Pacheco, J.~Ruiz-Ascencio, and J.~M. Rend{\'o}n-Mancha, ``Visual
  simultaneous localization and mapping: a survey,'' {\em Artificial
  Intelligence Review}, vol.~43, no.~1, pp.~55--81, 2015.

\bibitem{cole2006using}
D.~M. Cole and P.~M. Newman, ``Using laser range data for 3d slam in outdoor
  environments,'' in {\em International Conference on Robotics and Automation
  (ICRA)}, pp.~1556--1563, IEEE, 2006.

\bibitem{zhang2015visual}
J.~Zhang and S.~Singh, ``Visual-lidar odometry and mapping: Low-drift, robust,
  and fast,'' in {\em International Conference on Robotics and Automation
  (ICRA)}, pp.~2174--2181, IEEE, 2015.

\bibitem{nutzi2011fusion}
G.~N{\"u}tzi, S.~Weiss, D.~Scaramuzza, and R.~Siegwart, ``Fusion of imu and
  vision for absolute scale estimation in monocular slam,'' {\em Journal of
  intelligent \& robotic systems}, vol.~61, no.~1-4, pp.~287--299, 2011.

\bibitem{martinelli2012vision}
A.~Martinelli and R.~Siegwart, ``Vision and imu data fusion: Closed-form
  determination of the absolute scale, speed, and attitude,'' in {\em Handbook
  of Intelligent Vehicles}, pp.~1335--1354, Springer, 2012.

\bibitem{lupton2008removing}
T.~Lupton and S.~Sukkarieh, ``Removing scale biases and ambiguity from 6dof
  monocular slam using inertial,'' in {\em International Conference on Robotics
  and Automation (ICRA)}, pp.~3698--3703, IEEE, 2008.

\bibitem{grabe2013comparison}
V.~Grabe, H.~H. B{\"u}lthoff, and P.~R. Giordano, ``A comparison of scale
  estimation schemes for a quadrotor uav based on optical flow and imu
  measurements,'' in {\em International Conference on Intelligent Robots and
  Systems (IROS)}, pp.~5193--5200, IEEE, 2013.

\bibitem{mustaniemi2017inertial}
J.~Mustaniemi, J.~Kannala, S.~S{\"a}rkk{\"a}, J.~Matas, and J.~Heikkil{\"a},
  ``Inertial-based scale estimation for structure from motion on mobile
  devices,'' in {\em IEEE/RSJ International Conference on Intelligent Robots
  and Systems (IROS)}, pp.~4394--4401, IEEE, 2017.

\bibitem{xiong2017scale}
M.~Xiong, H.~Lu, D.~Xiong, J.~Xiao, and M.~Lv, ``Scale-aware monocular
  visual-inertial pose estimation for aerial robots,'' in {\em Chinese
  Automation Congress (CAC)}, pp.~7030--7034, IEEE, 2017.

\bibitem{tardif2010new}
J.-P. Tardif, M.~George, M.~Laverne, A.~Kelly, and A.~Stentz, ``A new approach
  to vision-aided inertial navigation,'' in {\em International Conference on
  Intelligent Robots and Systems (IROS)}, pp.~4161--4168, IEEE, 2010.

\bibitem{cvivsic2015stereo}
I.~Cvi{\v{s}}i{\'c} and I.~Petrovi{\'c}, ``Stereo odometry based on careful
  feature selection and tracking,'' in {\em European Conference on Mobile
  Robots}, pp.~1--6, IEEE, 2015.

\bibitem{buczko2016distinguish}
M.~Buczko and V.~Willert, ``How to distinguish inliers from outliers in visual
  odometry for high-speed automotive applications,'' in {\em IEEE Intelligent
  Vehicles Symposium (IV)}, pp.~478--483, IEEE, 2016.

\bibitem{persson2015robust}
M.~Persson, T.~Piccini, M.~Felsberg, and R.~Mester, ``Robust stereo visual
  odometry from monocular techniques,'' in {\em IEEE Intelligent Vehicles
  Symposium (IV)}, pp.~686--691, IEEE, 2015.

\bibitem{geiger2011stereoscan}
A.~Geiger, J.~Ziegler, and C.~Stiller, ``Stereoscan: Dense 3d reconstruction in
  real-time,'' in {\em IEEE Intelligent Vehicles Symposium (IV)}, pp.~963--968,
  IEEE, 2011.

\bibitem{choi2015new}
S.-I. Choi and S.-Y. Park, ``A new 2-point absolute pose estimation algorithm
  under planar motion,'' {\em Advanced Robotics}, vol.~29, no.~15,
  pp.~1005--1013, 2015.

\bibitem{scaramuzza2009absolute}
D.~Scaramuzza, F.~Fraundorfer, M.~Pollefeys, and R.~Siegwart, ``Absolute scale
  in structure from motion from a single vehicle mounted camera by exploiting
  nonholonomic constraints,'' in {\em International Conference on Computer
  Vision (ICCV)}, pp.~1413--1419, IEEE, 2009.

\bibitem{scaramuzza2008appearance}
D.~Scaramuzza and R.~Siegwart, ``Appearance-guided monocular omnidirectional
  visual odometry for outdoor ground vehicles,'' {\em IEEE Transactions on
  Robotics}, vol.~24, no.~5, pp.~1015--1026, 2008.

\bibitem{kitt2011monocular}
B.~M. Kitt, J.~Rehder, A.~D. Chambers, M.~Schonbein, H.~Lategahn, and S.~Singh,
  ``Monocular visual odometry using a planar road model to solve scale
  ambiguity,'' 2011.

\bibitem{lovegrove2011accurate}
S.~Lovegrove, A.~J. Davison, and J.~Ibanez-Guzm{\'a}n, ``Accurate visual
  odometry from a rear parking camera,'' in {\em IEEE Intelligent Vehicles
  Symposium (IV)}, pp.~788--793, IEEE, 2011.

\bibitem{grater2015robust}
J.~Grater, T.~Schwarze, and M.~Lauer, ``Robust scale estimation for monocular
  visual odometry using structure from motion and vanishing points,'' in {\em
  IEEE Intelligent Vehicles Symposium (IV)}, pp.~475--480, 2015.

\bibitem{gutierrez2012full}
D.~Guti{\'e}rrez-G{\'o}mez, L.~Puig, and J.~J. Guerrero, ``Full scaled 3d
  visual odometry from a single wearable omnidirectional camera,'' in {\em
  International Conference on Intelligent Robots and Systems (IROS)},
  pp.~4276--4281, IEEE, 2012.

\bibitem{pereira2017monocular}
F.~I. Pereira, G.~Ilha, J.~Luft, M.~Negreiros, and A.~Susin, ``Monocular visual
  odometry with cyclic estimation,'' in {\em SIBGRAPI Conference on Graphics,
  Patterns and Images (SIBGRAPI)}, pp.~1--6, IEEE, 2017.

\bibitem{fanani2017multimodal}
N.~Fanani, A.~St{\"u}rck, M.~Barnada, and R.~Mester, ``Multimodal scale
  estimation for monocular visual odometry,'' in {\em IEEE Intelligent Vehicles
  Symposium (IV)}, pp.~1714--1721, IEEE, 2017.

\bibitem{botterill2011bag}
T.~Botterill, S.~Mills, and R.~Green, ``Bag-of-words-driven, single-camera
  simultaneous localization and mapping,'' {\em Journal of Field Robotics},
  vol.~28, no.~2, pp.~204--226, 2011.

\bibitem{botterill2013correcting}
T.~Botterill, S.~Mills, and R.~Green, ``Correcting scale drift by object
  recognition in single-camera slam,'' {\em IEEE Transactions on Cybernetics},
  vol.~43, no.~6, pp.~1767--1780, 2013.

\bibitem{song2015high}
S.~Song, M.~Chandraker, and C.~Guest, ``High accuracy monocular sfm and scale
  correction for autonomous driving,'' {\em IEEE Transactions on Pattern
  Analysis and Machine Intelligence (TPAMI)}, 2015.

\bibitem{hilsenbeck2012scale}
S.~Hilsenbeck, A.~M{\"o}ller, R.~Huitl, G.~Schroth, M.~Kranz, and E.~Steinbach,
  ``Scale-preserving long-term visual odometry for indoor navigation,'' in {\em
  International Conference on Indoor Positioning and Indoor Navigation},
  pp.~1--10, IEEE, 2012.

\bibitem{lim2015monocular}
H.~Lim and S.~N. Sinha, ``Monocular localization of a moving person onboard a
  quadrotor mav,'' in {\em International Conference on Robotics and Automation
  (ICRA)}, pp.~2182--2189, IEEE, 2015.

\bibitem{sucar2017probabilistic}
E.~Sucar and J.-B. Hayet, ``Probabilistic global scale estimation for monoslam
  based on generic object detection,'' in {\em Computer Vision and Pattern
  Recognition Workshops (CVPRW)}, 2017.

\bibitem{mohanty2016deepvo}
V.~Mohanty, S.~Agrawal, S.~Datta, A.~Ghosh, V.~D. Sharma, and D.~Chakravarty,
  ``Deepvo: a deep learning approach for monocular visual odometry,'' {\em
  arXiv preprint arXiv:1611.06069}, 2016.

\bibitem{konda2015learning}
K.~R. Konda and R.~Memisevic, ``Learning visual odometry with a convolutional
  network.,'' in {\em VISAPP (1)}, pp.~486--490, 2015.

\bibitem{gomez2017learning}
R.~Gomez-Ojeda, Z.~Zhang, J.~Gonzalez-Jimenez, and D.~Scaramuzza,
  ``Learning-based image enhancement for visual odometry in challenging hdr
  environments,'' in {\em IEEE International Conference on Robotics and
  Automation (ICRA)}, pp.~805--811, May 2018.

\bibitem{wang2017end}
S.~Wang, R.~Clark, H.~Wen, and N.~Trigoni, ``End-to-end, sequence-to-sequence
  probabilistic visual odometry through deep neural networks,'' {\em
  International Journal of Robotics Research}, p.~0278364917734298, 2017.

\bibitem{costante2018ls}
G.~Costante and T.~A. Ciarfuglia, ``Ls-vo: Learning dense optical subspace for
  robust visual odometry estimation,'' {\em IEEE Robotics and Automation
  Letters}, vol.~3, no.~3, pp.~1735--1742, 2018.

\bibitem{zhao2018learning}
C.~Zhao, L.~Sun, P.~Purkait, T.~Duckett, and R.~Stolkin, ``Learning monocular
  visual odometry with dense 3d mapping from dense 3d flow,'' {\em arXiv
  preprint arXiv:1803.02286}, 2018.

\bibitem{zhan2018unsupervised}
H.~Zhan, R.~Garg, C.~Saroj~Weerasekera, K.~Li, H.~Agarwal, and I.~Reid,
  ``Unsupervised learning of monocular depth estimation and visual odometry
  with deep feature reconstruction,'' in {\em IEEE Conference on Computer
  Vision and Pattern Recognition (CVPR)}, June 2018.

\bibitem{zhou2016reliable}
D.~Zhou, Y.~Dai, and H.~Li, ``Reliable scale estimation and correction for
  monocular visual odometry,'' in {\em IEEE Intelligent Vehicles Symposium
  (IV)}, pp.~490--495, IEEE, 2016.

\bibitem{malis2007deeper}
E.~Malis and M.~Vargas, ``Deeper understanding of the homography decomposition
  for vision-based control,'' 2007.

\bibitem{faugeras1988motion}
O.~D. Faugeras and F.~Lustman, ``Motion and structure from motion in a
  piecewise planar environment,'' {\em International Journal of Pattern
  Recognition and Artificial Intelligence}, vol.~2, no.~03, pp.~485--508, 1988.

\bibitem{zhang19963d}
Z.~Zhang and A.~R. Hanson, ``3d reconstruction based on homography mapping,''
  {\em Proc. ARPA96}, pp.~1007--1012, 1996.

\bibitem{song2014robust}
S.~Song and M.~Chandraker, ``Robust scale estimation in real-time monocular sfm
  for autonomous driving,'' in {\em IEEE Conference on Computer Vision and
  Pattern Recognition (CVPR)}, pp.~1566--1573, 2014.

\bibitem{teichmann2018multinet}
M.~Teichmann, M.~Weber, M.~Zoellner, R.~Cipolla, and R.~Urtasun, ``Multinet:
  Real-time joint semantic reasoning for autonomous driving,'' in {\em IEEE
  Intelligent Vehicles Symposium (IV)}, pp.~1013--1020, IEEE, 2018.

\bibitem{lagarias1998convergence}
J.~C. Lagarias, J.~A. Reeds, M.~H. Wright, and P.~E. Wright, ``Convergence
  properties of the nelder--mead simplex method in low dimensions,'' {\em SIAM
  Journal on optimization}, vol.~9, no.~1, pp.~112--147, 1998.

\bibitem{frost2016object}
D.~P. Frost, D.~W. Murray, {\em et~al.}, ``Object-aware bundle adjustment for
  correcting monocular scale drift,'' in {\em International Conference on
  Robotics and Automation (ICRA)}, pp.~4770--4776, IEEE, 2016.

\bibitem{geiger2013vision}
A.~Geiger, P.~Lenz, C.~Stiller, and R.~Urtasun, ``Vision meets robotics: The
  kitti dataset,'' {\em International Journal of Robotics Research (IJRR)},
  2013.

\bibitem{klein2007parallel}
G.~Klein and D.~Murray, ``Parallel tracking and mapping for small ar
  workspaces,'' in {\em ACM International Symposium on Mixed and Augmented
  Reality (ISMAR)}, pp.~225--234, IEEE, 2007.

\bibitem{scaramuzza2006flexible}
D.~Scaramuzza, A.~Martinelli, and R.~Siegwart, ``A flexible technique for
  accurate omnidirectional camera calibration and structure from motion,'' in
  {\em IEEE International Conference on Computer Vision Systems}, pp.~45--45,
  2006.

\end{thebibliography}
	%
\begin{IEEEbiographynophoto}
{Dingfu Zhou} is currently a senior researcher at Robotics and Autonomous Driving Laboratory (RAL) of Baidu. Before joining in Baidu, he worked as a Post-Doc Researcher in the Research School of Engineering at the Australian National University, Canberra, Australia. He obtained his Ph.D degree in System and Control from Sorbonne Universit\'{e}s, Universit\'{e} de Technologie de Compi\`{e}gne, Compi\`{e}gne, France, in 2014. He received the B.E. degree and M.E degree both in signal and information processing from Northwestern Polytechnical University, Xian, China, in 2006, 2009, respectively. His research interests include Simultaneous Localization and Mapping, Structure from Motion, Classification and their application in Autonomous Driving.
\end{IEEEbiographynophoto}
\begin{IEEEbiographynophoto}
{Yuchao Dai} is a Professor with School of Electronics and Information at the Northwestern Polytechnical University (NPU). He received the B.E. degree, M.E. degree and Ph.D. degree all in signal and information processing from NPU, Xi’an, China, in 2005, 2008 and 2012, respectively. He was an ARC DECRA Fellow with the Research School of Engineering at the Australian National University, Canberra, Australia from 2014 to 2017 and a Research Fellow with the Research School of Computer Science at the Australian National University, Canberra, Australia from 2012 to 2014. His research interests include structure from motion, multi-view geometry, low-level computer vision, deep learning, compressive sensing and optimization. He won the Best Paper Award in IEEE CVPR 2012, the DSTO Best Fundamental Contribution to Image Processing Paper Prize at DICTA 2014, the Best Algorithm Prize in NRSFM Challenge at CVPR 2017, the Best Student Paper Prize at DICTA 2017 and the Best Deep/Machine Learning Paper Prize at APSIPA ASC 2017.
\end{IEEEbiographynophoto}
	\vspace{-1cm}
\begin{IEEEbiographynophoto}{Hongdong Li} is a Chief Investigator of the Australia ARC Centre of Excellence for Robotic Vision. He received the M.Sc. and PhD degrees in information and electronics engineering from Zhejiang University. Since 2004 he has joined the ANU as a postdoctoral fellow and also seconded to National ICT Australia (NICTA). His research interests include geometric computer vision, pattern recognition, computer graphics, and combinatorial optimization. Presently, he holds an Associate Professor position with ANU. He is an Associate Editor for IEEE T-PAMI, and served as Area Chair for recent CVPR, ICCV and ECCV.  He was a recipient of CVPR 2012 Best Paper Award, DSTO Best Fundamental Contribution to Image Processing Paper Prize in 2014 and best algorithm award in CVPR NRSFM Challenge 2017. He will be program co-chairing the ACCV 2018. 
\end{IEEEbiographynophoto}
	
	
	
	
\end{document}